%% file: preprint.tex
\documentclass[10pt,twocolumn,letterpaper]{article}

\usepackage[pagenumbers]{cvpr}      %

\input{preamble}

\definecolor{cvprblue}{rgb}{0.21,0.49,0.74}
\usepackage[pagebackref,breaklinks,colorlinks,citecolor=cvprblue]{hyperref}

\usepackage{colortbl}
\usepackage{pifont}

\newcommand{\ours}{MuM}
\newcommand{\dino}{DINOv3}
\newcommand{\croco}{CroCo~v2}

\def\paperID{2048} %
\def\confName{CVPR}
\def\confYear{2026}

\title{\ours: Multi-View Masked Image Modeling for 3D Vision}

\author{
David Nordström$^{1}$   
\and 
Johan Edstedt$^{2}$
\and
Fredrik Kahl$^{1}$    
\and
Georg Bökman$^{3}$%
\and 
\\
$^1$Chalmers University of Technology \hspace{2em} 
$^2$Linköping University
\hspace{2em} 
$^3$University of Amsterdam
}

\begin{document}
\maketitle

\begin{abstract}
Self-supervised learning on images seeks to extract meaningful visual representations from unlabeled data. When scaled to large datasets, this paradigm has achieved state-of-the-art performance and the resulting trained models such as DINOv3 have seen widespread adoption. However, most prior efforts are optimized for semantic understanding rather than geometric reasoning. One important exception is Cross-View Completion, CroCo, which is a form of masked autoencoding (MAE) tailored for 3D understanding. In this work, we continue on the path proposed by CroCo and focus on learning features tailored for 3D vision. In a nutshell, we extend MAE to arbitrarily many views of the same scene.
By uniformly masking all views and employing a lightweight decoder with inter-frame attention, our approach is inherently simpler and more scalable than CroCo.
We evaluate the resulting model, \ours, extensively on downstream tasks including 
feedforward reconstruction, dense image matching and relative pose estimation, finding that it outperforms the state-of-the-art visual encoders \dino~and \croco. Code is available at \href{https://github.com/davnords/mum}{https://github.com/davnords/mum}.
\end{abstract}

\section{Introduction} \label{sec:intro}

Self-supervised learning (SSL) enables large-scale pretraining without manual labels and has become a cornerstone of modern visual representation learning.
A popular family of SSL objectives for image data are based on masked autoencoding (MAE)~\cite{MAE:2021}.
In MAE, the training task consists of reconstructing a partially masked image.

Pixel reconstruction is an inherently low-level dense objective.
To increase 3D understanding, CroCo~\cite{croco} proposed conditioning the reconstruction with an unmasked reference view from the same scene.
However, this objective relies on substantial overlap between pairs, making the data sampling fragile, as pointed out by \citet{aligat0r:2025}.
To alleviate the sampling issue, \citet{aligat0r:2025} reformulated the task as a co-visibility segmentation objective.
On the other hand, relying on co-visibility necessitates access to ground-truth geometry, limiting its applicability for SSL.

Other approaches for SSL on images include the state-of-the-art DINO family \cite{dino, dinov2}, with DINOv3~\cite{simeoni2025dinov3} as its latest incarnation. DINOv3 produces rich image representations and is trained through self-distillation on billions of unlabeled images.

Recent works in 3D vision have successfully used DINOv2/3 networks as feature encoders~\cite{vggt:2025}, but the learned features in DINO are commonly described as semantic rather than geometric.
Furthermore, DINOv3 relies on carefully crafted heuristics such as Sinkhorn-Knopp centering to avoid training collapse and the requirement of a very large dataset makes it an inaccessible method for most of the academic community.

Motivated by the challenges outlined above, this work proposes a simple SSL objective for learning geometric features for 3D tasks while enabling flexible data sampling.
We show that extending the MAE objective to arbitrarily many views of the same scene yields a simple and powerful supervision strategy, surpassing \dino~in 3D vision pipelines while requiring roughly $30\times$ less training compute.\footnote{61,440 H100 hours (DINOv3-7B) vs.\ 4,608 A100 hours (\ours).}

\begin{figure*}[t]
  \centering
  \includegraphics[width=\textwidth]{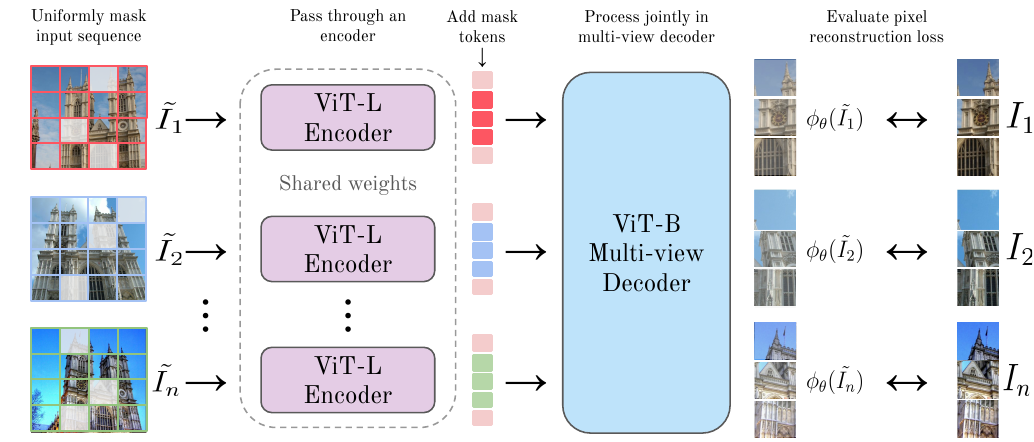}
  \caption{\textbf{\ours~ SSL pretraining}. An arbitrary number of input images from the same scene are first uniformly masked and processed independently in a ViT-L encoder. The representations are then jointly processed in a multi-view ViT-B decoder. The final representations are linearly mapped to pixel space and the reconstruction loss is computed according to Equation~\ref{eq:loss}.}
  \label{fig:architecture}
\end{figure*}

We call our objective, as well as the trained network, \ours~(\textbf{Mu}lti-View \textbf{M}asked Image Modeling). \ours~achieves strong downstream performance on a variety of tasks and is illustrated in Figure~\ref{fig:architecture}. The main contributions are as follows: 
\begin{enumerate}[topsep=0.5ex,itemsep=0ex,label=\arabic*.]
\item We extend masked image modeling to an arbitrary number of views of a scene and propose a simple, scalable training framework tailored for 3D vision.
\item We perform a comprehensive comparison of a wide range of models through dense feature matching performance using a linear probe.
\item We validate our approach on 3D vision tasks such as feedforward reconstruction, 
feature matching, and 
pose estimation by comparing \ours~as a feature extractor to the state-of-the-art baselines \dino~and \croco.
In particular, we train multiple feedforward reconstruction models inspired by, albeit smaller scale than, VGGT~\cite{vggt:2025}. Further, we train state-of-the-art two-view matchers following RoMa~\cite{roma:2023}.
\end{enumerate}

\newpage
\input{sec/2_related}
\input{sec/3_method}

\input{sec/4_results}
\input{sec/5_limitations}
\input{sec/6_conclusions}
\input{sec/acknowledgements}
{
    \small
    \bibliographystyle{ieeenat_fullname}
    \bibliography{main}
}
\appendix
\input{sec/X_suppl}

\end{document}

%% file: preamble.tex
\usepackage{amsmath}
\usepackage{amssymb}
\usepackage{amsfonts}
\usepackage{mathabx}

%% file: sec/2_related.tex
\section{Related work} \label{sec:related}

\paragraph{Masked Image Modeling (MIM).} The success of masked language modeling for self-supervised pre-training in NLP~\cite{BERT:2019, gpt2:2019, llama2} inspired similar efforts in computer vision. Following the popularization of ViTs~\cite{dosovitskiy2021an}, patch-based masking techniques were revisited to great success~\cite{MAE:2021, xie2022simmimsimpleframeworkmasked, multimae:2022, AIM:2024}. \citet{MAE:2021} introduced the masked auto-encoder (MAE) which uses a pixel reconstruction objective as its pre-text task.
The MAE reconstruction objective has been extended to, among other things, videos~\cite{tong2022videomae, wang2023videomaev2, carreira2025scaling4drepresentations} and multi-modal data~\cite{multimae:2022}.
MAE for videos has further been generalized to videos from multiple views~\cite{mv-worldmodels, mv2mae}, with similar motivations as our work.
Instead of videos, we focus on image data and, in particular, downstream applications involving unstructured and sparse sequences of images from a scene.

Closest to our work, \citet{croco} extended the MAE objective through cross-view completion (CroCo), and subsequently \croco~\cite{crocov2}, by conditioning on an unmasked reference view to encourage geometric understanding.
In this work, we also aim to learn features that are useful for 3D downstream tasks.
In contrast to CroCo, we sample an arbitrary number of frames with larger viewpoint variations and use alternating attention across them, yielding a simple, symmetric architecture that supports both single- and multi-frame training while offering greater robustness to frame sampling.

A different popular approach to SSL relies on aligning a student's features with those of a teacher. The most popular such approach relies on training the student and teacher from scratch jointly, often having the teacher be an exponential-moving-average of the student~\cite{meanteachers:2017}. Similar to MAE, one can use a masked reconstruction objective in this framework, but the reconstruction being in feature space rather than pixels~\cite{iBOT:2022, dinov2,CAPI:2025}. Combing this approach with instance discrimination~\cite{dino} has lead to the current state-of-the-art SSL objective, \dino~\cite{simeoni2025dinov3}. Although effective at producing highly semantic features, it is computationally demanding, as it relies on a student-teacher setup and attends between masked patches. In this work, we study how this objective transfers to geometric downstream tasks and, interestingly, conclude that a simple pixel reconstruction objective works better at learning geometric features. We demonstrate the effectiveness of our objective on a range of 3D computer vision applications.

\begin{figure*}[t]
  \centering
  \begin{subfigure}[t]{0.32\textwidth}
    \centering
    \includegraphics[width=\textwidth]{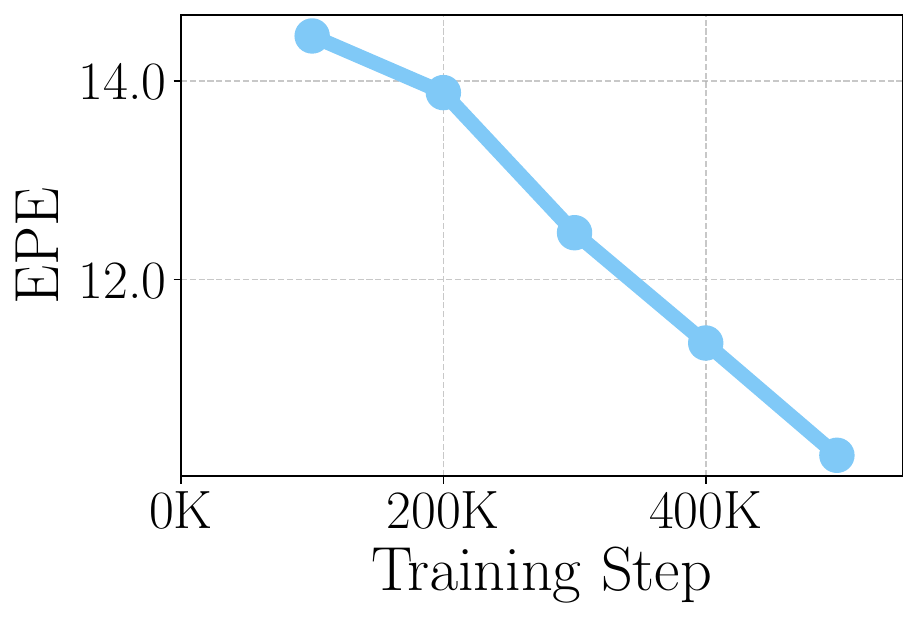}
    \caption{MegaDepth linear~$\downarrow$}
  \end{subfigure}
  \hfill
  \begin{subfigure}[t]{0.32\textwidth}
    \centering
    \includegraphics[width=\textwidth]{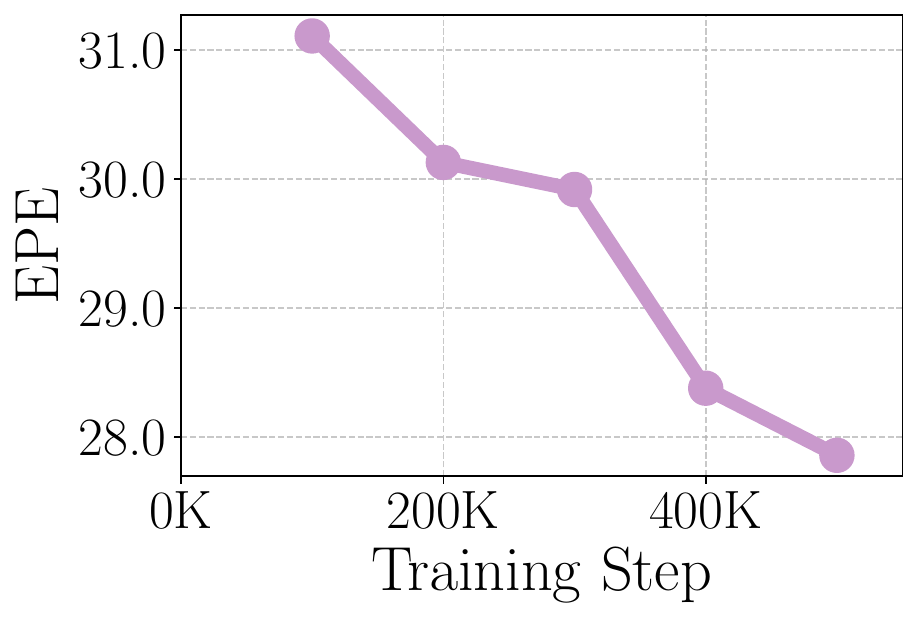}
    \caption{ScanNet linear~$\downarrow$}
  \end{subfigure}
  \hfill
  \begin{subfigure}[t]{0.32\textwidth}
    \centering
    \includegraphics[width=\textwidth]{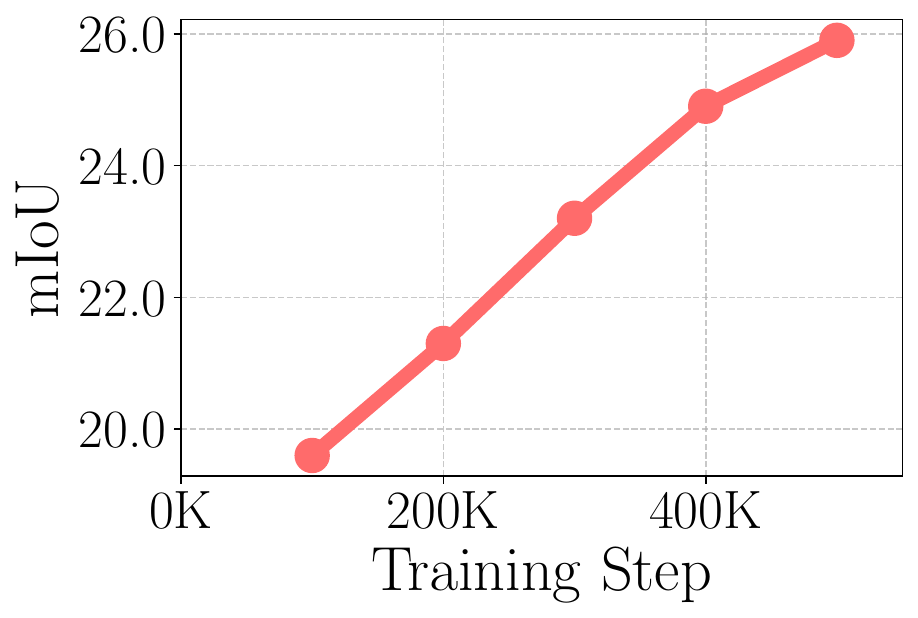}
    \caption{ADE20K $k$-NN~$\uparrow$}
  \end{subfigure}

  \caption{\textbf{Training dynamics.} Dense matching with a linear probe and semantic segmentation performance during 500K steps of training. }
  \label{fig:training-dynamics}
\end{figure*}

\paragraph{Multiple-view Geometry.} In an effort to replace handcrafted methods for multiple-view geometry~\cite{Hartley_Zisserman_2004} with learning-based approaches, two-view neural network architectures have been used in image matching~\cite{loftr:2021, edstedt2022dkmdensekernelizedfeature, lightglue:2023, roma:2023}, pose estimation~\cite{reloc3r:2025} and stereo reconstruction~\cite{dust3r:2024, mast3r:2024}. To extend its usefulness in traditional 3D computer vision tasks such as structure-from-motion (SfM), architectures processing arbitrarily many frames jointly were introduced~\cite{brynte-cvpr-2024, mvdust3r:2024, spann3r:2024, fast3r:2025, must3r:2025, vggt:2025, yugay2025visualodometrytransformers, keetha2025mapanything}. Recently, transformer architectures trained on large-scale 3D-annotated datasets have been proposed to jointly predict all scene attributes in a single forward pass, notable examples being VGGT~\cite{vggt:2025} and MapAnything~\cite{keetha2025mapanything}. VGGT allows inter-frame communication through alternating frame-wise and global attention. In this work, we adopt the same attention mechanism between frames in our decoder but in the context of SSL. We further propose a symmetric architecture, rather than anchoring a reference view, similar to \citet{wang2025pi3}.

%% file: sec/3_method.tex
\section{Method}

We begin by formalizing multi-view masked image modeling (Section~\ref{subsec:prob}), followed by introducing our proposed learning objective (Section~\ref{subsec:obj}), architecture (Section~\ref{subsec:arch}), and training procedure (Section~\ref{subsec:training}).

\subsection{Problem definition} \label{subsec:prob}

Standard masked image modeling considers reconstructing an image from a partially masked version of itself. We adopt a more general formulation that naturally extends this idea to multiple views.

Let $\mathcal{I} = (I_1, I_2, \dots, I_n)$ be a sequence of $n$ images of the same scene, each divided into $N$ non-overlapping patches. For each view $i$, we mask a subset of the patches according to binary masks $M_i\in\{0,1\}^N$, with masking ratios $\gamma_i \in [0, 1]$.
We interpret $M_i$ as being $1$ for the masked patches and
obtain a partially masked sequence of images $$ \tilde{I}_i = \widebar{M_i}\odot I_i,
\quad \widebar{M_i}=1-M_i,
$$
where $\odot$ denotes element-wise patch selection.

The learning objective is to train a neural network $\phi_\theta$ to predict a target representation $f(\mathcal{I})$ from the visible patches in $\tilde{\mathcal{I}} =
(\tilde{I}_1, \tilde{I}_2, \dots, \tilde{I}_n)$. 
The loss (up to a normalizing constant) is given by:
\begin{equation} \label{eq:loss}
    \mathcal{L}(\theta)=
    \sum_{i=1}^n %
    ||M_i\odot (\phi_\theta(\tilde{I_i})-f(I_i))||^2.
\end{equation}
The role of $f$ is to specify the reconstruction target. In the simplest case, $f$ is the identity (possibly with normalization), yielding pixel-space regression as in MAE or CroCo. More generally, $f$ may produce higher level representations, which shows the close connection to self-distilled teacher neworks, as in iBOT, DINOv2, and CAPI~\cite{CAPI:2025}, as well as frozen pretrained teachers, as in BeiT~\cite{beit:2022} and EVA~\cite{eva:2022}.   

The original MAE objective corresponds to the single-view setting ($n=1$) and a fixed masking ratio $\gamma_1=0.75$ %
whereas the cross-view completion formulation in CroCo uses two views ($n=2$) with $\gamma_1=0.9$ and $\gamma_2=0$.

\subsection{Learning objective}\label{subsec:obj}
We design \ours~by uniformly sampling sequences between $n=2$ and $n=24$ frames and using a uniform masking ratio of $\gamma_i=0.75$ for all $i$. The target representation $f$ simply normalizes the pixel values in a patch by its mean and standard deviation.
Our objective is therefore exactly equal to MAE for single views ($n=1$).
In this way, we address the difficulties in CroCo of sampling co-visible pairs as there is always a fallback to the standard MAE objective.
Furthermore, our objective straightforwardly generalizes to an arbitrary number of views, while it is not obvious how to generalize CroCo to more than two views in terms of selecting the masking ratios $\gamma_i$.
We ablate our design choices and the objective in Section~\ref{subsec:ablations}. 

\subsection{Network architecture} \label{subsec:arch}

We implement $\phi_\theta$ as a Vision Transformer (ViT)~\cite{dosovitskiy2021an} with an encoder-decoder structure (cf.\ Figure~\ref{fig:architecture}), similar to MAE and CroCo. For the ViT, we use modern architectural choices such as axial RoPE~\cite{rope, simeoni2025dinov3}. Each input image $I_i\in\mathcal{I}$ is patchified into $N$ tokens.
Following Section~\ref{subsec:prob}, a fraction $\gamma_i$ of the patches is removed, and the remaining tokens are passed through a ViT encoder. Thereafter, learnable mask tokens are appended for missing patches. These tokens are then processed jointly in a ViT-B decoder using $L=6$ alternating attention blocks.
Each block performs (i) frame-wise attention, restricting attention within each view, followed by (ii) global attention, allowing tokens to attend across all views.
This symmetric design avoids designating any reference frame.
Lastly, the token embeddings for each image are passed through a linear prediction head that regresses the normalized RGB values for each patch and the loss is computed by Equation~\ref{eq:loss}.

\subsection{Training}\label{subsec:training}

\paragraph{Implementation Details.} We train \ours~ for 500k steps by minimizing the training loss (\ref{eq:loss}) using the AdamW~\cite{adamw:2019} optimizer. We use a cosine learning rate scheduler and a linear warmup of 25K steps. Following the linear scaling rule of \citet{goyal2018accuratelargeminibatchsgd}, $\text{lr} = \frac{\text{blr}}{256} \times \text{batch size}$, we use a base learning rate of $1\times10^{-4}$, yielding a peak rate of $2.4\times10^{-3}$ for our global batch size of 6144. Similar to VGGT, for each batch we randomly select a sequence length uniformly between 2 and 24. Based on this length, we then sample as many training scenes as possible without exceeding 96 total frames per GPU. We resize the images to $256\times 256$ and randomly apply horizontal mirroring augmentation per frame. Pre-training on $64\times$A100 GPUs takes roughly three days. As shown in Figure~\ref{fig:training-dynamics}, the evaluation metrics consistently improve during training.

\begin{figure}[t]
  \centering
  \includegraphics[width=\columnwidth]{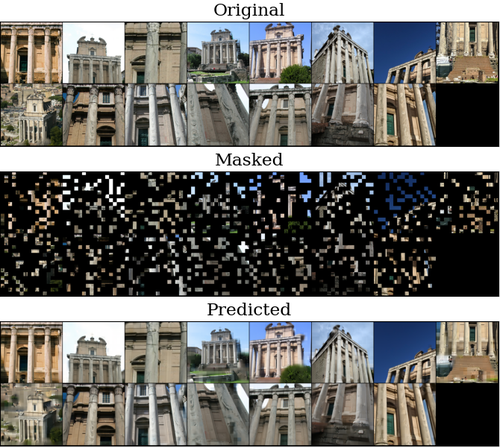}
  \caption{\textbf{Data example}. We illustrate a sequence sampled from the data and the predicted reconstructions. Note, as we train with a normalized pixel objective, we visualize the prediction plus the patch mean. Our data samples generally feature difficult viewpoint changes and varying co-visibility. Details about our sampling technique are provided in Appendix~\ref{appendix:pretrain}. }
  \label{fig:data-example}
\end{figure}

\paragraph{Training Data.} We train on a diverse set of 3D datasets inspired by the mix used in DUSt3R~\cite{dust3r:2024} and VGGT. In particular, we train on 3DStreetView~\cite{streetview:2016}, ARKitScenes~\cite{dehghan2021arkitscenes},  BlendedMVS~\cite{yao2020blendedmvs}, CO3D~\cite{co3d:2021}, DL3DV10K~\cite{ling2024dl3dv}, Hypersim~\cite{hypersim:2021}, MegaDepth~\cite{MegaDepthLi18}, MegaScenes~\cite{tung2024megascenes}, ScanNet++~\cite{scannetpp:2023}, RealEstate10K~\cite{realestate10k}, and WildRGBD~\cite{wilrgbd:2024}. When applicable, we utilize only the training splits. In total, the collection comprises around 20 million individual frames (see Table~\ref{tab:dataset_mix} for a detailed breakdown). In Figure~\ref{fig:data-example}, we visualize a representative sequence sampled from our data collection. \ours~provides a simple symmetric objective which enables us to mix multi-view and single-view data. Inspired by \citet{simeoni2025dinov3}, we use a 10\% probability to sample data from ImageNet-1K~\cite{imageNet2009, russakovsky2015imagenet}. Another advantage of our architecture is that it is robust to simpler sampling schemes and does not strictly require covisibility.

\begin{table}[h] \centering
\small
\caption{\textbf{Dataset mixture} and sample sizes for {\ours } pre-training.}
\label{tab:dataset_mix}
\setlength{\tabcolsep}{2pt}
\begin{tabular}{lccr}
\hline
\specialrule{1.5pt}{0pt}{0pt} 
Datasets                          & Type            & Prob. & Frames \\ 
\specialrule{1.5pt}{0.5pt}{0.5pt} 
3DStreetView~\cite{streetview:2016}  & Outdoor / Real       &10\%&  6282k \\
WildRGBD~\cite{wilrgbd:2024}  & Objects / Real       &10\%&  3352k \\
DL3DV10K~\cite{ling2024dl3dv}  & Outdoor / Real       &10\%&  3317k \\
RealEstate10K~\cite{realestate10k}  & Indoor / Real       &10\% &  1332k \\
ImageNet1K\footnotemark~\cite{imageNet2009, russakovsky2015imagenet} & Objects / Real  & 10\% &  1281k \\
MegaScenes~\cite{tung2024megascenes}    & Outdoor / Real        &10\%&  1235k \\
ScanNet++~\cite{scannetpp:2023}  & Indoor / Real       &10\%&  1045k \\
CO3D~\cite{co3d:2021}  & Objects / Real       &5\%&  546k \\
MegaDepth~\cite{MegaDepthLi18}    & Outdoor / Real        &10\%&  205k \\
BlendedMVS~\cite{yao2020blendedmvs}  & Aerial / Synthetic       &5\%&  115k \\
ARKitScenes~\cite{dehghan2021arkitscenes}& Indoor / Real          &5\%&  113k \\
Hypersim~\cite{hypersim:2021}& Indoor / Synthetic          &5\%&  43k \\
\specialrule{1.5pt}{0.5pt}{0.5pt} 
\textbf{Total}                        &                     & & \textbf{19462k} \\
\specialrule{1.5pt}{0pt}{0pt} 
\end{tabular}
\normalsize
\end{table}
\footnotetext{ImageNet1K consists of only single-view data.}

%% file: sec/4_results.tex
\section{Results}

We evaluate \ours\ on a broad range of downstream tasks and compare primarily to the state-of-the-art feature encoders DINOv3 and \croco. For computationally lighter evaluations, we also compare to MAE. Unless otherwise specified, we use the ViT-L backbone, which is the standard choice in recent 3D vision pipelines, \eg, VGGT, RoMa~\cite{roma:2023}, MapAnything, UFM~\cite{zhang2025ufm}.

We organize the evaluations by the number of input views $n$ used at inference time, distinguishing between multi-view tasks ($n > 2)$, two-view tasks ($n = 2)$, and single-view tasks ($n = 1)$.

For multi-view tasks ($n>2$), we evaluate scene understanding by feedforward 3D reconstruction. We consider both the case with a frozen backbone and finetuning (Section~\ref{subsec:reconstruction}).
For two-view tasks ($n=2$), we study dense matching (Section~\ref{subsec:matching}) and relative pose estimation (Section~\ref{subsec:relpose}).
For single-view tasks ($n=1$), we assess monocular depth estimation (Section~\ref{subsec:mono-depth}), surface normal estimation (Section~\ref{subsec:surface-normals}), and semantic recognition tasks such as image classification and segmentation (Section~\ref{subsec:semantic}).
Finally, we conduct an ablation study on the objective and design choices (Section~\ref{subsec:ablations}).

Across all multi-view 3D tasks ($n > 1$), \ours\ consistently outperforms DINOv3 and \croco. Notably, it surpasses DINOv3 despite using substantially less compute and training data. On monocular tasks such as image classification and segmentation, which favor stronger semantic inductive biases, \ours~performs below DINOv3 but remains superior to \croco. Code and weights will be released on GitHub.

\subsection{Multi-view tasks ($n > 2$): Feedforward 3D reconstruction} \label{subsec:reconstruction}

We investigate 3D understanding by multi-view feedforward reconstruction. The task is to train a neural network to estimate the 3D attributes of a scene, including camera parameters, point maps, and depth maps, from an arbitrary number of views.

\paragraph{Frozen encoder.} Inspired by VGGT and MapAnything, we design an evaluation protocol where we train a multi-view ViT on-top of a frozen encoder. Following \citet{vggt:2025}, we employ a ViT camera head and separate DPT~\cite{ranftl2021vision} heads for depths and points. The model has around 700M parameters whereof around half are trainable. We train for 100K steps on BlendedMVS~\cite{yao2020blendedmvs}, CO3Dv2~\cite{co3d:2021}, Hypersim~\cite{hypersim:2021}, MegaDepth~\cite{MegaDepthLi18}, MVS-Synth~\cite{DeepMVS}, PointOdyssey~\cite{zheng2023point}, ScanNet~\cite{dai2017scannet}, VKitti~\cite{cabon2020vkitti2}, and WildRGBD~\cite{wilrgbd:2024} randomly sampling between 2 and 24 frames per scene. Each training run takes approximately 3 days on $32\times$H200. More details can be found in Appendix~\ref{appendix:VGGT}.    

Following~\citet{vggt:2025}, we evaluate multi-view camera pose estimation on  Re10K~\cite{realestate10k} and CO3Dv2~\cite{co3d:2021} and point cloud estimation on DTU~\cite{dtu:2014} and ETH3D~\cite{eth3d:2017}. In addition, we evaluate on MegaDepth~\cite{MegaDepthLi18}. The evaluated scenes have not been seen during training.  The results show the usefulness of our frozen features for 3D reconstruction. \ours~improves performance compared to \dino~and \croco~on five different datasets and two different tasks, showing its versatility on both indoor, outdoor and object centric scenes.

\paragraph{Distillation finetuning.} The full \ours~architecture enables rapid finetuning on multi-view tasks, similar to how \croco~can be adapted to binocular tasks. To illustrate this, we keep the decoder and simply append camera, depth and point heads to the full model. We train by distilling VGGT on our pretraining data using an unweighted L2 loss:   
\begin{equation}
    \mathcal{L}(\theta)
    =\sum_i^{n} \| \mathcal{P}_t-\mathcal{P}_s\|^2 + \| \mathcal{C}_t-\mathcal{C}_s\|^2 +\| \mathcal{D}_t-\mathcal{D}_s\|^2
    \label{eq:distill}
\end{equation}
where we sum the differences between the student and teacher over the frames and $\mathcal{P}$, $\mathcal{C}$, and $\mathcal{D}$ represent world points, camera parameters, and depth maps, respectively.

We train for 100K steps using a learning rate of $1\times 10^{-4}$, taking roughly 3 days on $16\times$H200. More details can be found in Appendix~\ref{appendix:experimental-details}. As shown in Table~\ref{tab:camera-pose} and \ref{tab:multi-view-estimation}, \ours~provides significant advantages over random (rand.) initialization.

\begin{table}[t]
\centering
     \caption{
\textbf{Multi-view camera pose estimation.} Evaluating AUC@30 ($\uparrow$) for 10 random frames.
\label{tab:camera-pose}} \centering
    \begin{tabular}{@{}lrrrr@{}}
        \toprule
        Method & CO3Dv2 & Re10K & MegaDepth  \\
        \midrule
        \multicolumn{4}{@{}l@{}}{\small \textit{Frozen encoder}} \\
        \croco & 58.2 & 27.7 & 60.7 \\
        DINOv3 & 66.9 & 36.7 & 59.3 \\
        
        \ours & \bfseries 71.5 & \bfseries 50.8  & \bfseries 73.0 \\
        \midrule
        \multicolumn{4}{@{}l@{}}{\small \textit{Distillation finetuning}} \\
        Rand. & 10.6 & 28.0 & 41.8 \\
        \ours & \bfseries 62.6 & \bfseries 50.9 & \bfseries 78.6 \\
        \bottomrule
        \end{tabular}
\end{table}

\begin{table}[t]
\centering
     \caption{
\textbf{Pointcloud estimation}. Reporting median accuracy and completeness on DTU and ETH3D.
\label{tab:multi-view-estimation}} \centering
    \begin{tabular}{@{}l r rr r rr @{}}
        \toprule
        Method && \multicolumn{2}{@{}c@{}}{DTU} && \multicolumn{2}{@{}c@{}}{ETH3D}  \\
        \cmidrule{3-4} \cmidrule{6-7}
        && Acc.~$\downarrow$ & Comp.~$\downarrow$ && Acc.~$\downarrow$ & Comp.~$\downarrow$\\
        \midrule
        \multicolumn{7}{@{}l@{}}{\small \textit{Frozen encoder}} \\
        \croco && 8.5 & 12.3  && 0.9 & 1.0\\
        DINOv3 && 6.4 & 3.7 && 0.9 & 1.0  \\
        \ours  && \bfseries 3.7 & \bfseries 1.6 && \bfseries 0.8 & \bfseries 0.8  \\
        \midrule
        \multicolumn{7}{@{}l@{}}{\small \textit{Distillation finetuning}} \\
        Rand. && 8.4 & 11.7 && 1.0 & 1.4 \\
        \ours && \bfseries 6.4 & \bfseries 2.6 && \bfseries 0.6 & \bfseries 0.5 \\
        \bottomrule
        \end{tabular}
\end{table}

\subsection{Two-view tasks ($n = 2$): Matching} \label{subsec:matching}

We now study frozen features for image matching. We perform both a lightweight evaluation with a linear probe and full RoMa training~\cite{roma:2023}.    

\paragraph{Linear probe.} We train a linear probe on-top of the frozen features followed by a kernel nearest neighbor matcher. We evaluate average end-point-error (EPE) and robustness ($\mathcal{R}$), where robustness is defined as the share
of matches with an error less than 32 pixels. In Table~\ref{tab:matchbench}, we perform a comprehensive comparison of a wide range of vision backbones. We find that \ours~produces the best representations for dense image matching. In particular, \ours~outperforms both \croco~and \dino. Notably, the performance of DINOv2 improves significantly after being fine-tuned in VGGT, indicating the need for a geometric objective. We found the results to be stable regardless of whether layer normalization is included. In Table~\ref{tab:cosine-matching} in the Appendix, we further evaluate matching without probing and find degraded performance for pixel-level objectives.

\begin{table}[t]
\centering
     \caption{
\textbf{Comparison on dense feature matching} through linear probing. End-point-error (EPE) and robustness ($\mathcal{R}$) on MegaDepth-1500~\cite{MegaDepthLi18, loftr:2021} and ScanNet-1500~\cite{dai2017scannet, sarlin2020superglue}.
\label{tab:matchbench}
}
\small 
  \centering
    \begin{tabular}{@{}llr@{} rr rr@{\ }}
        \toprule 
        Method & Arch. &&
        \multicolumn{2}{@{}c@{}}{MegaDepth} &
        \multicolumn{2}{@{}c@{}}{ScanNet}\\
        \cmidrule{4-5} \cmidrule{6-7} &&& EPE~$\downarrow$ & $\mathcal{R}\uparrow$ & EPE~$\downarrow$ & $\mathcal{R}\uparrow$ \\
        \midrule
        \multicolumn{7}{@{}l@{}}{\small \textit{Weakly-supervised}} \\
        SigLIP 2~\cite{siglip2} & ViT-G/16  && 62.1 & 42.3  & 75.8 & 34.6 \\
        PEcore~\cite{bolya2025PerceptionEncoder} & ViT-G/14 && 75.8 & 28.8 & 93.6 & 23.2 \\
        \midrule
        \multicolumn{7}{@{}l@{}}{\small \textit{Instance discrimination}} \\
        iBOT~\cite{iBOT:2022} & ViT-L/16  && 28.5 & 73.7 & 38.6 & 66.1 \\
        DINOv2~\cite{dinov2} & ViT-L/14  && 26.5 & 77.0 & 43.9 & 60.7 \\
        DINOv2$^1$ & ViT-L/14  && 19.3 & 84.6 & 29.7 & 74.4 \\
        DINOv3~\cite{simeoni2025dinov3} & ViT-L/16 && 19.0 & 86.4 & 28.7 & 78.0 \\
        DINOv3~\cite{simeoni2025dinov3} & ViT-G/16 && 17.1 & 87.2 & \bfseries 27.3 & 77.2 \\
        \midrule
        \multicolumn{7}{@{}l@{}}{\small \textit{Masked image modeling}} \\
        CAPI~\cite{CAPI:2025} & ViT-L/14  && 21.4 & 82.2 & 32.9 & 71.9\\
        I-JEPA~\cite{ijepa:2023} & ViT-H/14  && 42.6 & 58.4 & 74.7 & 36.0 \\
        V-JEPA 2~\cite{assran2025vjepa2} & ViT-H/16  && 60.6 & 37.8 & 86.5 & 23.0 \\
        MAE~\cite{MAE:2021} & ViT-L/16  && 29.7 & 73.4 & 35.0 & 71.6 \\
        \croco~\cite{crocov2} & ViT-L/16  && 27.3 & 75.7 & 39.0 & 66.6\\
        \croco$^2$ & ViT-L/16  && 22.0 & 80.9 & 29.0 & 76.5\\
        \midrule
        \ours$^3_{32\times\text{A100}}$ & ViT-L/16 && 12.0 & 93.7 & 30.2 & 76.0 \\
        \ours$_{64\times\text{A100}}$ & ViT-L/16 && \bfseries 10.2 & \bfseries 94.2 & 27.9 & \bfseries 78.9 \\
        \bottomrule		
        \multicolumn{7}{l}{\footnotesize $^1$Weights when finetuned in VGGT~\cite{vggt:2025}.} \\
        \multicolumn{7}{l}{\footnotesize $^2$Weights when finetuned in DUSt3R~\cite{dust3r:2024}.} \\
        \multicolumn{7}{l}{\footnotesize $^3$A lighter training run using 66 A100 days vs. 96 of \croco.} \\
      \end{tabular}
\end{table}

\begin{table}
    \centering
    \caption{\textbf{Comparison on RoMa.} Dense matching measured in 100-percentage correct keypoints (PCK) (lower is better).}
    \begin{tabular}{@{}l rrrr}
    \toprule
         Method $\downarrow$\quad\quad\quad 100-PCK$@$ $\rightarrow$       & 1px  & 3px  & 5px   \\
         \midrule
         \multicolumn{4}{@{}l@{}}{\small MegaDepth-1500~\citep{MegaDepthLi18,sun2021loftr}} \\
     \textcolor{gray}{RoMa (DINOv2)}$^\dagger$ & \textcolor{gray}{13.3} & \textcolor{gray}{4.6} & \textcolor{gray}{3.1} \\
     \croco  &	13.0 & 4.6 & 3.2 \\
     DINOv3  &	13.8 & 5.2 & 3.8 \\
     \ours  &	\bfseries 12.5 & \bfseries 4.0 & \bfseries 2.6 \\
      \midrule
        \multicolumn{4}{@{}l@{}}{\small ScanNet-1500~\citep{dai2017scannet,sarlin2020superglue}} \\ 
        \textcolor{gray}{RoMa (DINOv2)}$^\dagger$ & \textcolor{gray}{92.1} & \textcolor{gray}{60.4} & \textcolor{gray}{38.3} \\
        \croco  &	92.3 & 61.6 & 40.0 \\
     DINOv3  &	 92.3 & 61.0 & 38.9 \\
     \ours  &	\bfseries 92.2 & \bfseries 60.5 & \bfseries 38.1 \\
         \bottomrule
         \multicolumn{4}{l}{\footnotesize $^\dagger$Not trained by us and patch size 14 instead of 16.} \\
    \end{tabular}

    \label{tab:roma}
\end{table}

\begin{table*}
    \centering
    \caption{\textbf{Two-view relative pose estimation.} Comparing AUC$@$ for different encoders through finetuning.}
    \begin{tabular}{@{}l rrrr r rrr r rrr}
    \toprule
        Method & \multicolumn{3}{@{}c@{}}{MegaDepth~$\uparrow$} && \multicolumn{3}{@{}c@{}}{Re10K~$\uparrow$} && \multicolumn{3}{@{}c@{}}{BlendedMVS~$\uparrow$} \\ 
         \cmidrule{2-4} \cmidrule{6-8} \cmidrule{10-12} & 5$^\circ$  & 10$^\circ$  & 20$^\circ$ &&  5$^\circ$  & 10$^\circ$  & 20$^\circ$ &&  5$^\circ$  & 10$^\circ$  & 20$^\circ$    \\
         \midrule
        \croco  & 13.9 & 30.0 & 48.0 && 8.6 & 24.0 & \bfseries 44.9 && \bfseries 7.7 & 18.3 & 32.1 \\
     \dino  & 15.6 & 32.5 & 51.2 && 7.3 & 20.7 & 39.6 && 3.3 & 10.3 & 23.0 \\
     \midrule
     \ours  & \bfseries 26.7 & \bfseries 47.0 & \bfseries 65.0 && \bfseries 10.3 & \bfseries 25.3 & 44.5 && 6.6 & \bfseries 18.4 & \bfseries 35.8 \\
         \bottomrule
    \end{tabular}

    \label{tab:rel-pose}
\end{table*}

\paragraph{RoMa.} RoMa is a state-of-the-art dense matcher achieving robust results under challenging viewpoint conditions. The model utilizes a frozen coarse encoder followed by a match decoder and convolutional refiners. We simply replace the frozen coarse encoder in RoMa and compare different backbones. We train on MegaDepth using the official pipeline, taking approximately 5 days on $4\times$A100-80GB. We also evaluate the outdoor model on the indoor dataset ScanNet. We report the performance in Table~\ref{tab:roma}. \ours~outperforms \dino~as well as \croco~on both datasets. 

\subsection{Two-view tasks ($n = 2$): Relative pose} \label{subsec:relpose}
We now consider the task of relative pose estimation between two cameras through finetuning. Inspired by~\citet{reloc3r:2025}, we extract image tokens through a siamese encoder and the tokens are then jointly processed through cross-attention in the decoder. The features are then fed to a relative camera pose regression network and a motion averaging module and trained through a translation and rotation loss. We use a randomly initialized ViT for the decoder and compare different encoders. We jointly train the encoder and decoder on pairs extracted from BlendedMVS, CO3Dv2, DL3DV10K, MegaDepth, MVS-Synth, RealEstate10K, ScanNet, VKitti, and WildRGBD and report the results in Table~\ref{tab:rel-pose}. Training takes roughly a day on $32\times$A100. Further details can be found in Appendix~\ref{appendix:experimental-details}.

The task gives \croco~a slight structural advantage, as its decoder also uses binocular cross-attention and pairs are sampled similarly to its training data. Nevertheless, \ours~performs comparably or better and outperforms DINOv3.

\subsection{Single-view tasks ($n = 1$): Depth estimation}\label{subsec:mono-depth}

We now consider the task of monocular depth estimation through linear probing on top of the frozen features. We adopt the evaluation protocol of \citet{dinov2}~and evaluate on NYUd~\cite{NYUd:2012} and KITTI~\cite{KITTI:2017}, reporting the performance in Table~\ref{tab:depth}. We find that \ours~outperforms \croco~and MAE but achieves worse results than DINOv3. 

\begin{table}[t]
\centering
     \caption{
\textbf{Depth estimation with frozen features.} We report performance when training a linear
classifier on top of the frozen features. We report the RMSE metric on the 2 datasets.
\label{tab:depth}
}
  \centering
    \begin{tabular}{@{}lrrrr@{}}
        \toprule 
        Method &&
        NYUd~$\downarrow$ && KITTI~$\downarrow$ \\
        \midrule
        MAE && 0.59 && 4.19  \\
        \croco && 0.51 && 4.09  \\
        DINOv3 && \bfseries 0.34 && \bfseries 2.63 \\
        \midrule
        \ours && 0.41 && 3.89 \\
        \bottomrule		
      \end{tabular}
\end{table}

\subsection{Single-view tasks ($n = 1$): Surface normals}\label{subsec:surface-normals}
 
We follow Probe3D~\cite{banani2024probing3dawarenessvisual} to evaluate surface normals with a DPT probe trained for 10 epochs on NYUv2~\cite{NYUd:2012}. We report the performance on the test set in ~\Cref{tab:surface-normal}. As with monocular depth estimation, \ours~outperforms \croco~and MAE but trails DINOv3.

\begin{table}[t]
\centering
     \caption{
\textbf{Surface normal estimation}. We estimate surface normals with a DPT probe, following Probe3D~\cite{banani2024probing3dawarenessvisual}, and report percentage recall at different angular thresholds and RMSE. 
\label{tab:surface-normal}} \centering
    \begin{tabular}{@{}lrrrr@{}}
        \toprule
        Method & 11.25$^\circ$~$\uparrow$ & 22.5$^\circ$~$\uparrow$ & 30$^\circ$~$\uparrow$ & RMSE~$\downarrow$  \\
        \midrule
        MAE & 49.1 & 72.0 & 79.8 & 26.1 \\
        CroCo v2 & 50.0 & 71.7 & 79.3 & 26.8 \\
        DINOv3 & \bfseries 60.4 & \bfseries 79.3 & \bfseries 85.4 & \bfseries 22.5 \\
        \midrule
        \ours & 54.3 & 75.8 & 82.6 & 24.5 \\
        \bottomrule
        \end{tabular}
\end{table}

\subsection{Single-view tasks ($n = 1$): Semantic recognition} \label{subsec:semantic}

While our objective is to create a strong feature encoder for 3D vision tasks, we evaluate its performance on the high-level semantic tasks of image classification on ImageNet-1K and semantic segmentation on ADE20K~\cite{ade20k:2017, ade20k:2019}. We closely follow the evaluation protocol of~\citet{CAPI:2025}. For classification, we use an \textit{attentive probe}~\citep{ijepa:2023}. For segmentation, we use two lightweight classifiers on top of frozen local features, logistic regression and $k$-NN. We get robust results by doing an extensive sweep over hyperparameters and report the results in Table~\ref{tab:semantic}. DINOv3 leads on semantic tasks due to its instance segmentation (DINO) loss, which encourages semantic understanding. \ours~surpasses \croco~and MAE in semantic segmentation, though for classification, it slightly lags behind MAE that is exclusively trained on ImageNet-1K.

\begin{table}[t]
\centering
     \caption{
\textbf{Semantic tasks.} We train an attentive probe on frozen features for classification (IN1K, accuracy), and evaluate semantic segmentation (ADE20K, mIoU) using logistic regression and $k$-NN. The performance of the pixel reconstruction objectives lag behind the highly semantic DINOv2 and v3 models.
\label{tab:semantic}
}
  \centering
    \begin{tabular}{@{}lrrrrr@{\ }}
        \toprule 
        Method &&
        IN1K && \multicolumn{2}{@{}c@{}}{ADE20K} \\
        \cmidrule{3-3} \cmidrule{5-6}
         && Att. && Lin. & $k$-NN  \\
        \midrule
        \multicolumn{6}{@{}l@{}}{\small \textit{Latent prediction}} \\
        CAPI && 82.9 && 34.4 & 29.2 \\
        DINOv2 && 85.4 && 39.0 & 34.8 \\
        DINOv3 && \bfseries 86.9 && \bfseries 43.1 & \bfseries 41.5\\
        \midrule
        \multicolumn{6}{@{}l@{}}{\small \textit{Pixel reconstruction}} \\
        MAE && 76.6 && 27.6 & 21.5 \\
        CroCo v2 && 57.4 && 25.0 & 18.8 \\
        \midrule
        \ours && 70.8 && 30.2 & 26.0\\
        \bottomrule		
      \end{tabular}
\end{table}

\subsection{Ablation studies} \label{subsec:ablations}

\paragraph{SSL objective.} We verify the choice of extending MAE to multi-view by comparing to the three most popular SSL objectives for 3D vision: DINOv2, CroCo v2 and MAE. We also considered CAPI, but found the large batch size to be prohibitive. We make a comparison in an equal data and training setting by training a ViT-B for 100K steps on MegaDepth with 128 images per GPU. We report the linear probing matching performance and training time in Table~\ref{tab:mv-loss}. 

\begin{table}[t]
\centering
     \caption{
\textbf{SSL objective ablation.} We evaluate different objectives by training each on MegaDepth for 100K steps and report linear probing matching performance. Reformulating MAE to multiple views is fast and gives robust matches. Time denotes the number of hours of training on an 8$\times$A100 node. 
\label{tab:mv-loss}
}
  \centering
    \begin{tabular}{@{}p{0.55\linewidth}rr@{\ }}
        \toprule 
        Method & Time~$\downarrow$ &
        EPE~$\downarrow$ 
        \\
        \midrule
        DINOv2$^\dagger$ & 38h & 28.9 
        \\
        \hspace{0.5em}w/ multi-view & 38h & 28.4 
        \\
        \croco & \bfseries 12h &
        41.4 
        \\
        MAE &  \bfseries 12h & 18.7 
        \\
        \rowcolor{green!25} \hspace{0.5em}w/ multi-view & \bfseries 12h & \bfseries 12.5 
        \\
        \bottomrule		
        \multicolumn{3}{l}{\footnotesize $^\dagger$We use the SimDINOv2 objective~\cite{wu2025simdino} for multi-view stability.} \\
      \end{tabular}
\end{table}

\definecolor{baselinecolor}{gray}{.9}
\newcommand{\baseline}[1]{\cellcolor{baselinecolor}{#1}}
\begin{table*}[t]
  \centering
  \small{
    \begin{tabular}[b]{ccc}
      \begin{subtable}[t]{0.3\textwidth}
        \centering
        \begin{tabular}{lrr}
          & EPE~$\downarrow$ & Acc.~$\uparrow$  \\
          \midrule
          2, 6 &  12.8 & 47.4  \\
          2, 12 & 12.5 & 47.4  \\
          2, 24 & \baseline{\bfseries 10.6} & \baseline{\bfseries 47.8}  \\
        \end{tabular}
        \caption{\textbf{Sequence length.} Longer sequence lengths improve matching performance.}
      \end{subtable}
      &
  \begin{subtable}[t]{0.3\textwidth}
        \centering
        \begin{tabular}{lrr}
          & EPE~$\downarrow$ & Acc.~$\uparrow$  \\
          \midrule
          65\% & 13.3 & \bfseries 49.1  \\
          75\% & \baseline{\bfseries 10.6} & \baseline{47.8} \\
          85\% & 12.7 & 45.6  \\
        \end{tabular}
        \caption{\textbf{Mask ratio.} The same masking ratio as MAE (75\%) works well.}
      \end{subtable}
          &
      \begin{subtable}[t]{0.3\textwidth}
        \centering
        \begin{tabular}{lrr}
          & EPE~$\downarrow$ & Acc.~$\uparrow$  \\
          \midrule
          w/ ref.  & 11.9 & 47.7  \\
          w/o ref. & \baseline{\bfseries 10.6} & \baseline{\bfseries 47.8} \\
        \end{tabular}
        \caption{\textbf{Reference view.} An unmasked reference view is not beneficial.}
      \end{subtable}
      \\
      \\
    \begin{subtable}[t]{0.3\textwidth}
        \centering
        \begin{tabular}{lrr}
          & EPE~$\downarrow$ & Acc.~$\uparrow$  \\
          \midrule
          Encoder & 16.7 & 43.6  \\
          Decoder & \baseline{\bfseries 10.6} & \baseline{\bfseries 47.8}  \\
        \end{tabular}
        \caption{\textbf{Frame communication.} Alternating attention works when used in the decoder.}
      \end{subtable}
      & 
            \begin{subtable}[t]{0.3\textwidth}
        \centering
        \begin{tabular}{lrr}
          & EPE~$\downarrow$ & Acc.~$\uparrow$  \\
          \midrule
          \rowcolor{white}
          Absolute & 12.1 & 46.9  \\
          RoPE & \baseline{\bfseries 10.6} & \baseline{\bfseries 47.8}  \\
        \end{tabular}
        \caption{\textbf{Positional embedding}. A modern axial RoPE is superior to learned embeddings.}
      \end{subtable}
    &
      \begin{subtable}[t]{0.3\textwidth}
        \centering
        \begin{tabular}{lrr}
          & EPE~$\downarrow$ & Acc.~$\uparrow$  \\
          \midrule
          w/o norm & 13.4 & 45.9  \\
          w/ norm & \baseline{\bfseries 10.6} & \baseline{\bfseries 47.8}  \\
        \end{tabular}
        \caption{\textbf{Reconstruction objective.} A normalized pixel objective is effective.}
      \end{subtable}

          \\
      \\

\end{tabular}

  }
  \caption{\textbf{Ablations.} We train ViT-B/16 on MegaDepth for 100K steps on $8\times$A100 with 128 frames per GPU and report the linear probing performance on dense matching on MegaDepth and image classification on ImageNet-1K. Default settings are marked in \colorbox{baselinecolor}{gray}.}
  \label{tab:ablations}
\end{table*}

We attempt to reformulate the monocular objectives in a multi-view context by randomly sampling between 2 and 12 images with alternating attention blocks providing inter-frame communication. The change constrains batch sampling to include multiple frames from the same scene and introduces information sharing between those frames, effectively reducing the variation in the data and limiting the number of scenes seen during training. Yet, we find that, while only providing a minute increase to the DINOv2 objective, it boosts MAE performance significantly. The MAE objective is substantially simpler, both conceptually and computationally, showcased through more than $3\times$speedup in training. We also experimented with other reconstruction targets than pixels but, while improving semantic performance, we experienced worse geometric performance.  

Compared to a simple single-view MAE objective, \croco~performs poorly. This can also be seen from Table~\ref{tab:matchbench}, where MAE achieves similar results to \croco~ on MegaDepth and ScanNet while the latter has trained on similar data. As \croco~uses a higher masking ratio (90\%), we tried training on our sampled pairs (generally harder) and the pairs provided in the original paper. We find the latter to produce better results and report those.

\paragraph{Architecture.} To verify the design of our multi-view architecture, we ablate the main design choices in Table~\ref{tab:ablations}. We find that many previously motivated architectural improvements are also relevant in this setting. Such as the 75\% masking ratio and normalized objective used by MAE, RoPE and inter-frame communication in the decoder used by CroCo v2 and uniform sampling between 2 and 24 frames used by VGGT. In contrast, an unmasked reference view, as is used in CroCo v2, slightly degrades performance while complicating the architecture. 

\begin{figure}[t]
  \centering
  \includegraphics[width=.8\columnwidth]{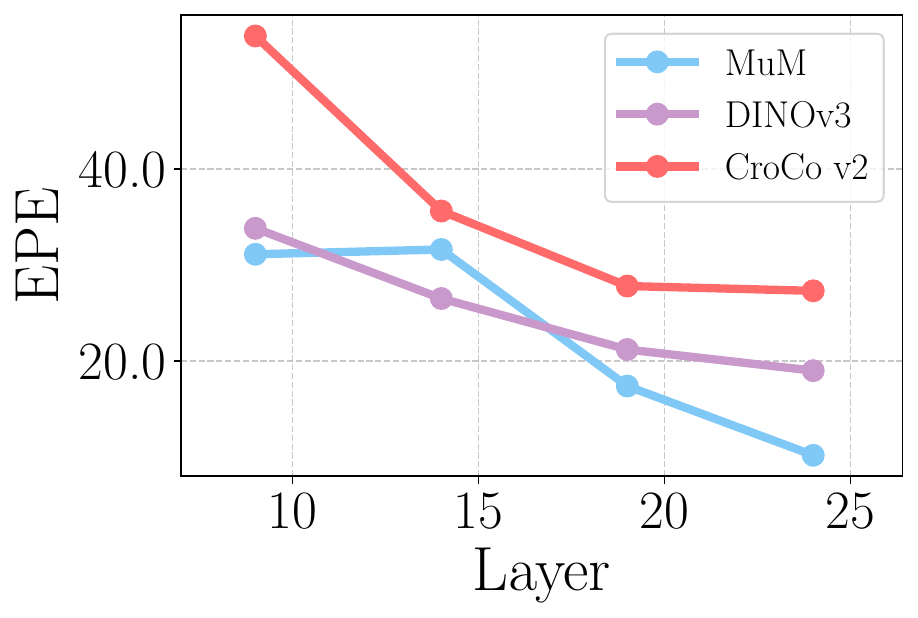}
  \caption{\textbf{Feature layer}. We report EPE for linear probing on MegaDepth. The later layers provide better representations.}
  \label{fig:feature-layers}
\end{figure}

\paragraph{Feature layers.} Finally, we validate our choice of extracting feature representations from the final layer throughout the preceding sections. We use ViT-L with 24 layers and report linear probing dense matching performance in Figure~\ref{fig:feature-layers} after extracting features from different layers. We find that performance improves with deeper features.

%% file: sec/5_limitations.tex
\section{Limitations}
In~\Cref{subsec:ablations}, we found the simple pixel reconstruction loss to outperform baselines such as DINOv2 for training on multi-view data, in particular per GPU hour.
Our findings there do not preclude the possibility of future research on self-distillation training on multi-view data to prove fruitful and
we consider it an interesting direction to try to combine the semantic performance of DINO with geometric information from multi-view data.
Moreover, due to computational constraints, we are unable to scale pretraining further than presented in the paper.
Similarly, we do not have the resources to fully replicate the training and hence performance of recent feedforward reconstruction methods such as VGGT and MapAnything.
However, we believe that our results provide a strong indication that our model would perform well if incorporated into such 3D vision pipelines and hope to inspire further research on the topic.

%% file: sec/6_conclusions.tex
\section{Conclusions}

We introduce a multi-view masked-image-modeling objective that extends MAE to an arbitrary number of views of the same scene. 
The objective enables learning meaningful features for 3D tasks in a self-supervised manner.
We showcase its effectiveness by training a ViT-L and evaluating the resulting model, \ours, on downstream tasks such as feedforward reconstruction, feature matching, and relative pose estimation.
While trained using substantially less compute and data than DINOv3, \ours~gives improved results when used as a feature extractor in 3D vision tasks. 
\newpage

%% file: sec/acknowledgements.tex
\section*{Acknowledgements}
This work was supported by the Wallenberg Artificial
Intelligence, Autonomous Systems and Software Program
(WASP), funded by the Knut and Alice Wallenberg Foundation, and by the strategic research environment ELLIIT, funded by the Swedish government. 
The computational resources were provided by the
National Academic Infrastructure for Supercomputing in
Sweden (NAISS) at C3SE, partially funded by the Swedish Research
Council through grant agreement no.~2022-06725, and by
the Berzelius resource, provided by the Knut and Alice Wallenberg Foundation at the National Supercomputer Centre.

%% file: sec/X_suppl.tex
\clearpage
\setcounter{page}{1}
\maketitlesupplementary

\section{Further Details on Pretraining}\label{appendix:pretrain}

\paragraph{Data sequences. } Below, we outline the selection of sequences of frames. In the following section, we detail how to sample batches from sequences. For information on the dataset mixture, we refer to Section~\ref{subsec:training} of the main manuscript. Our protocol for finding sequences is simple, as strict co-visibility is not required. This greatly simplifies things and allows us to, without much effort, scale to around 20 million individual frames. MegaDepth and MegaScenes are the only datasets where we employ a more complicated sampling scheme. This is because each scene of the dataset is highly diverse and has no natural temporal sequencing. 

For the former, we use the image overlap scores computed by D2-Net~\cite{d2net:2019} and greedily sample 1,000 sequences per scene by randomly selecting an anchor frame and then chaining together images based on positive overlap. For the latter, we compute the IoU heuristic similar to CroCo and iteratively select the next image that maximizes 3D point overlap with the current image while maintaining sufficient camera viewpoint diversity. Starting from images with the most visible 3D points, we greedily build sequences until no suitable next image can be found, discarding sequences shorter than 24. 

For datasets where each scene is small (\eg, CO3D, DL3DV10K, Hypersim, WildRGBD), we simply select all frames in the scene as sequences and randomly sample from those during training. For the remaining datasets (3DStreetView, RealEstate10K, ScanNet++, BlendedMVS, and ARKitScenes), we use simple heuristics to form sequences. Since the frames per scene are numerically ordered, where adjacent frames are also spatially adjacent in the scene, we chunk them. We find that a chunk size of 100 frames usually works fine. After forming sequences of frames, we now detail how to select the frames in a batch. 

\begin{figure}[t]
  \centering
  \includegraphics[width=.8\columnwidth]{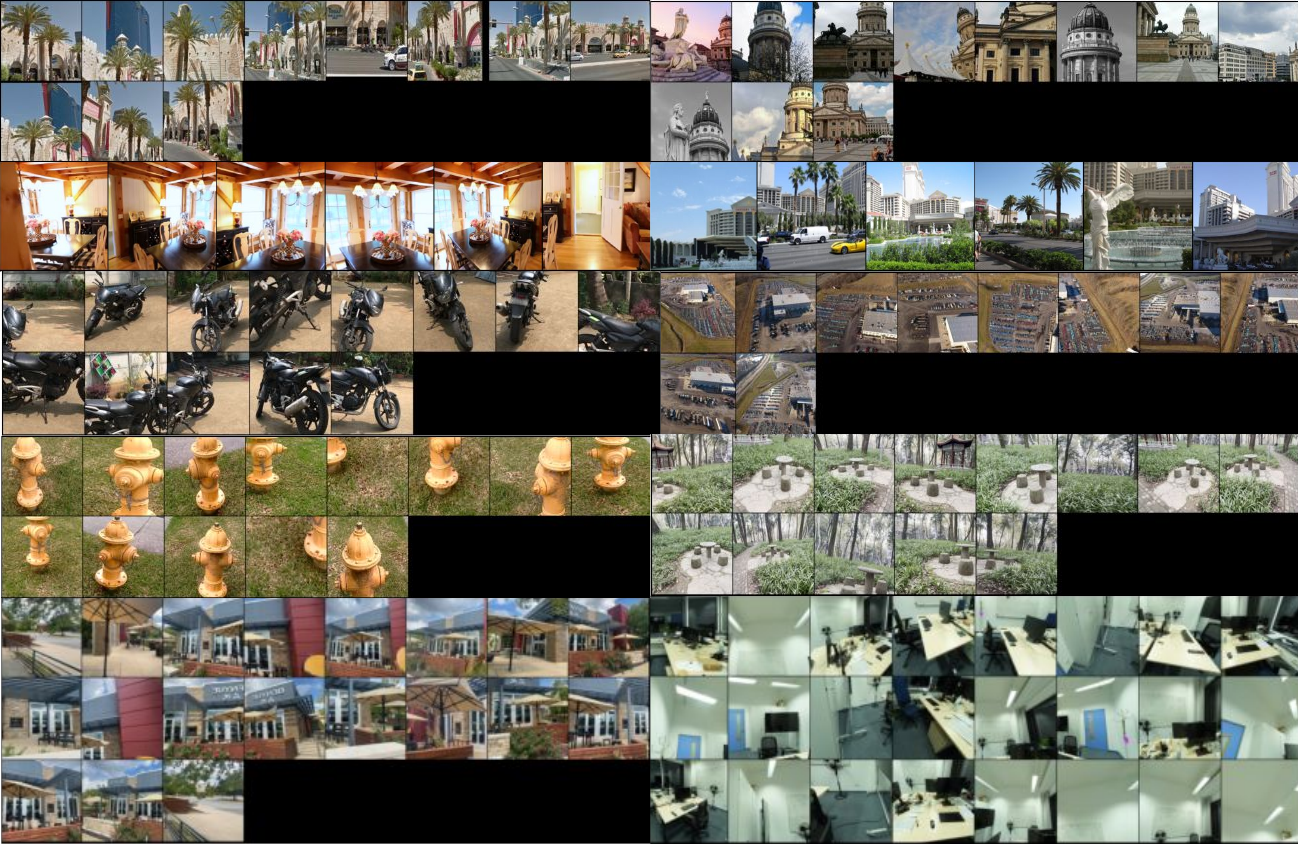}
  \caption{\textbf{Data examples}. We visualize random sampled sequences from the dataset. Sequence lengths vary.}
  \label{fig:data-examples}
\end{figure}

\paragraph{Sampling. } For each batch, independently on each GPU, we either (i) sample a full batch from ImageNet-1K (10\% probability) or (ii) sample multi-view data (90\% probability). To sample multi-view data, we uniformly sample a sequence length $S$ between 2 and 24 frames. Given the sequence length, we then iteratively sample datasets with the weighting given in Table~\ref{tab:dataset_mix} and uniformly select a sequence therein. See Figure~\ref{fig:data-examples} for qualitative examples of sequences. We stop when the total number of frames per GPU is about to exceed 96. This gives us a batch shape of $(\text{floor}(\frac{96}{S}), S,3,256,256 )$. This is done independently per GPU. We sometimes refer to 96 as the batch size and use this in the linear scaling rule for learning rate~\cite{goyal2018accuratelargeminibatchsgd}.   

\paragraph{Architecture. } The total model is made up of a ViT-L encoder (width 1024, depth 24 and 16 heads) followed by a ViT-B multi-view decoder (width 768, depth 12 and 12 heads). This is the exact same size as the final CroCo v2 model and thus also DUSt3R and MASt3R (excluding heads). Half of the Transformer blocks in the multi-view decoder have a global attention window and half only attend to patches within the same frame. The first block is a frame-wise attention block. The total number of trainable parameters in the model is 389.5 million, whereof around 302.3 million come from the encoder. The final layer, after the multi-view decoder, is a linear layer $W\in \mathbb{R}^{768\times768}$. The dimensions can be understood from the way that it maps each patch between the 768 embedding dimensions of the Transformer decoder to the $3\times16\times16=768$ color values in a patch. The loss is then computed by comparing the output of this linear mapping to the normalized (mean subtracted and divided by the standard deviation) pixel values in the ground truth patch. 

\paragraph{Compute Resources.} \label{appendix:compute-resources} In Table~\ref{tab:gpu-hours}, we detail the computing resources used for the main experiments. Failed or discarded training runs are excluded. The evaluations not included required only a linear probe or a $k$-NN classifier. These were run on a single A100 and took no longer than a couple of hours. The reporting is conducted towards academic transparency in the compute resources required to replicate our results.

\begin{table}[htbp]
  \centering
    \caption{
    \textbf{GPU hours.}  We report the compute used for the main experiments. We omit linear probing and $k$-NN validations.
  }
  \label{tab:gpu-hours}
  \small{
   \begin{tabular}{@{}lllrr}
        \toprule 
        Experiment & Model & GPUs & Time \\
        \midrule
        Pretraining & \ours$^\dagger$ & $32\times$A100 & 3 days \\
         Pretraining& \ours & $64\times$A100 & 3 days \\
        \midrule
        RoMa & \ours & $4\times$A100 & 5 days \\
        RoMa & DINOv3 & $4\times$A100 & 5 days \\
        RoMa & CroCov2 & $4\times$A100 & 5 days \\
        \midrule
        Rel. pose & \ours & $32\times$A100 & 1.5 days \\
        Rel. pose & DINOv3 & $32\times$A100 & 1.5 days \\
        Rel. pose & CroCov2 & $32\times$A100 & 1.5 days \\
        \midrule
        3D recon. (frozen) & \ours & $32\times$H200 & 3 days \\
        3D recon. (frozen) & DINOv3 & $32\times$H200 & 3 days \\
        3D recon. (frozen) & CroCov2 & $32\times$H200 & 3 days \\
        3D recon. (finetune) & \ours & $16\times$H200 & 3 days \\
        3D recon. (finetune) & Rand. & $16\times$H200 & 3 days \\
        \bottomrule
        \multicolumn{4}{l}{\footnotesize $^\dagger$A lighter training run using 66 A100 days vs. 96 of CroCov2.} \\
    \end{tabular}
  }
\end{table}

\section{Further Details on Experiments}\label{appendix:experimental-details}

\paragraph{Feedforward 3D reconstruction.}\label{appendix:VGGT} We take inspiration from the design of VGGT and use the same heads, excluding the tracking head. Namely, we use DPT heads for world points and depths and an iterative ViT for the camera head. For the frozen encoder experiments, the points and depth heads also output confidence maps, we train for 100K steps with a learning rate of $2\times10^{-4}$ on $32\times$H200 for around 3 days, and sample data following VGGT, namely, using 48 frames per GPU and an image resolution of 512. We use the AdamW~\cite{adamw:2019} optimizer and minimize the same training loss as in VGGT:
\begin{equation}
    \mathcal{L}(\theta)=\mathcal{L}_{\text{camera}}+\mathcal{L}_{\text{depth}}+\mathcal{L}_{\text{pmap}}  
\end{equation}
Where the camera loss is given by $\mathcal{L}_{\text{camera}}=\sum_{i=1}^{N}||\hat{c}_i-c_i||_{\mathcal{H}}$ where $c_i$ are the ground truth camera parameters and $\hat{c}_i$ are the predictions in frame $i$ and $||\cdot||_{\mathcal{H}}$ denotes the Huber loss. The depth objective, $\mathcal{L}_{\text{depth}}$, follows DUSt3R and uses an aleatoric-uncertainty loss. We further incorporate the gradient-based term used in VGGT. The pointmap loss, $\mathcal{L}_{\text{pmap}}$, is similarly defined.   

The finetuning distillation protocol is slightly lighter using half the batch size (due to training on half the number of GPUs) and with a learning rate of $1\times10^{-4}$. The distillation loss is simply the sum over the L2 errors (given in Equation~\ref{eq:distill}). We train without confidence weights.

\paragraph{Dense matching.} For the linear probing evaluation we take as basis the MegaDepth1500 and ScanNet1500 dense evaluation protocols in RoMa and DKM. We train a square linear projection on top of the frozen encoder features followed by a log-softmax kernel nearest neighbor matcher to estimate the dense warp. We achieve a robust metric by sweeping over learning rates and weight decays in parallel. We train for 25K steps with batch size 8 and evaluate every 2,500 steps, reporting the best performance achieved. The evaluation protocol takes roughly an hour on an A100.

For RoMa, we follow their exact training setup. which is the following. Training for 3 million steps on $560\times560$ resolution with a frozen encoder. This takes roughly 5 days on $4\times$A100-80GB. We opt to only train on MegaDepth (commonly referred to as outdoor) and use this model for all evaluations. %

\paragraph{Two-view relative pose estimation.}\label{appendix:reloc3r} The decoder architecture follows exactly Reloc3r~\cite{reloc3r:2025} but is not initialized from DUSt3R but instead trained from scratch. We train the decoder and the heads with a learning rate of $2\times10^{-4}$ and finetune the encoder with a learning rate of $2\times10^{-5}$. We use a total batch size of 1024 and train on $16\times$A100 for approximately 1 day.

\paragraph{Monocular depth estimation.}\label{appendix:monocular-depth} 

We follow the protocol in DINOv2~\cite{dinov2}. Namely, we use the MMCV~\cite{mmcv} package and apply random cropping, rotation and flipping augmentation. We crop the images to $480\times 640$ and treat it as a classification problem with 256 bins. We use a combination of cross entropy and gradient loss. We train with AdamW, cosine annealing and a peak learning rate of $1\times 10^{-4}$.

\paragraph{Semantic recognition.}\label{appendix:semantic}

For both classification and segmentation, we follow exactly the protocol in CAPI~\cite{CAPI:2025}. In particular, the classification protocol uses an attentive probe. This is due to \ours, CroCo v2 and MAE, not having a meaningful global representation in the same way as the DINO family (following its instance discrimination objective on the CLS token). The attentive probe is trained in a supervised fashion and its output is processed by a trained linear layer.

\section{Additional experiments}

\paragraph{Qualitative examples.} 

\begin{figure}[t]
  \centering
  \includegraphics[width=.8\columnwidth]{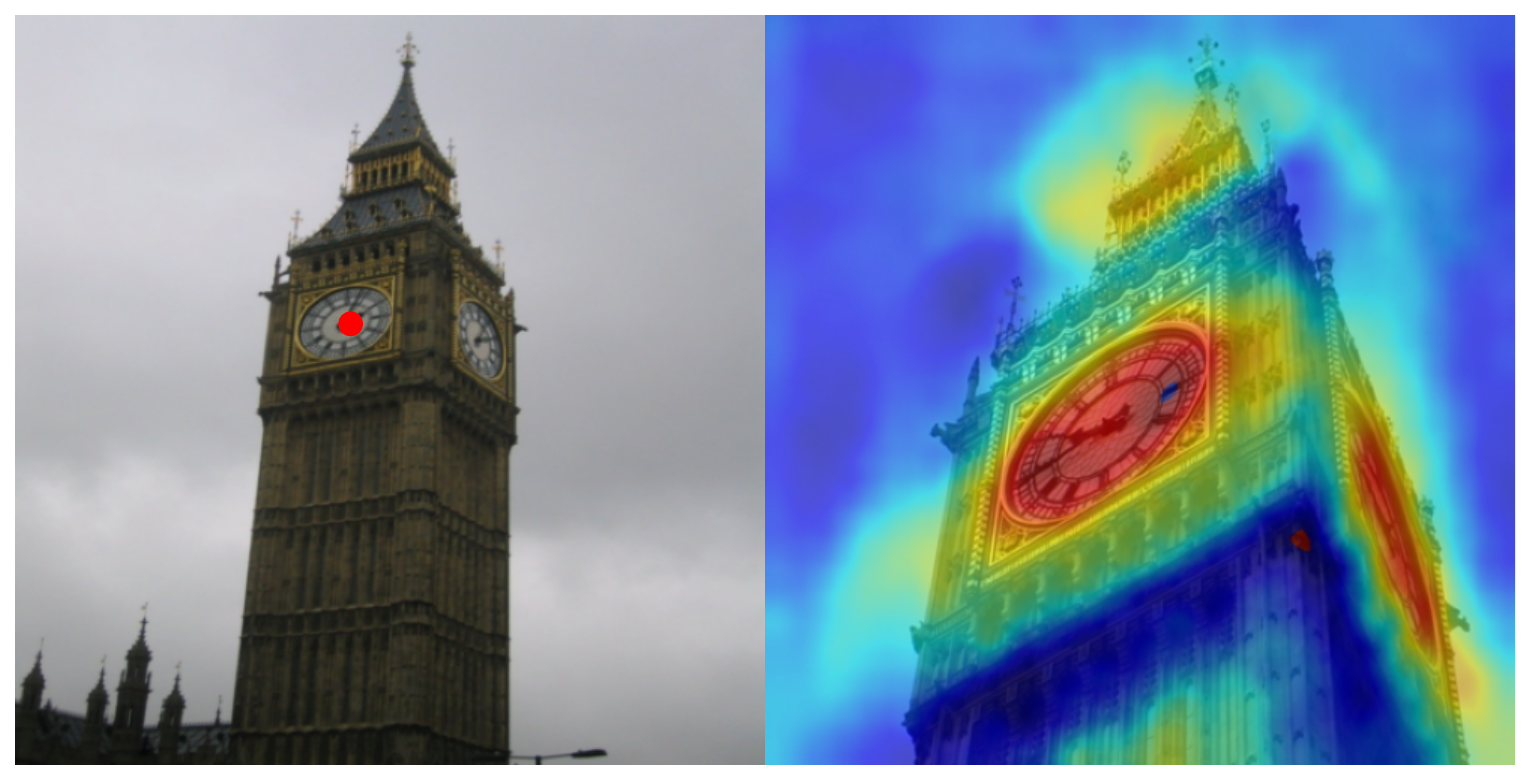}
  \caption{\textbf{Attention map for query patch}. We visualize the global attention map in the decoder for a query patch. We find that the attention score is highest for matching patches.}
  \label{fig:attention-match}
\end{figure}

Inspired by ZeroCo~\cite{zeroco:cvpr}, we visualize the global attention map between a query patch in the first image and all the patches in the second image. The result, illustrated in Figure~\ref{fig:attention-match}, clearly illustrates that the attention map contains information about correspondences between the two frames. An interesting future direction is to investigate zero-shot matching directly from the attention map, similar to ZeroCo~\cite{zeroco:cvpr}.

\begin{figure}[t]
  \centering
  \includegraphics[width=.8\columnwidth]{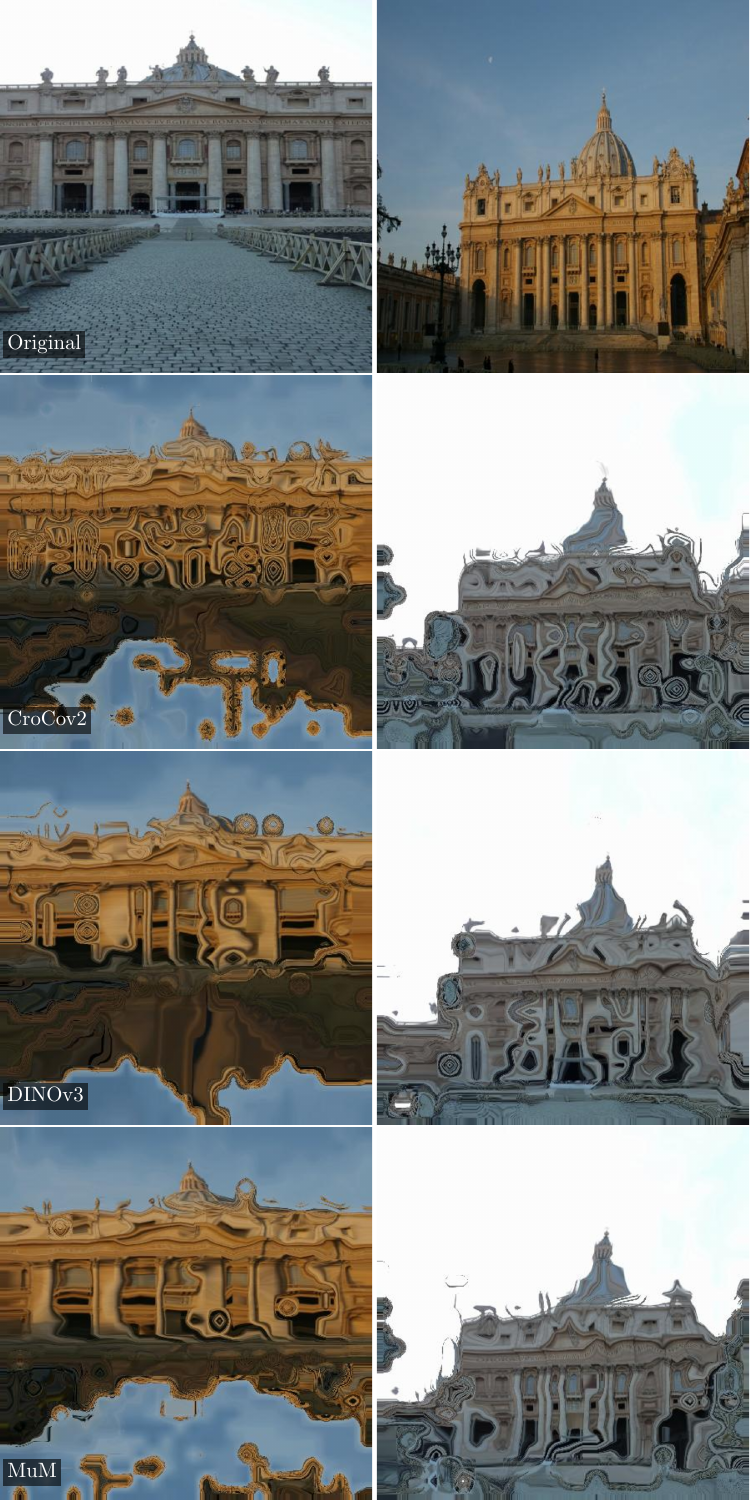}
  \caption{\textbf{Dense warp}. Warp estimated through linear probing on MegaDepth on top of frozen features.}
  \label{fig:dense-warp}
\end{figure}

We also visualize the dense warp estimated through linear probing in Section~\ref{subsec:matching} in Figure~\ref{fig:dense-warp}. 

\paragraph{Cosine similarity feature matching.} In our main results, we report the linear probing dense feature matching accuracy. However, prior work has explored the matchability of the raw features. While \citet{banani2024probing3dawarenessvisual} notes that more expressive probes are necessary to determine 3D understanding, they opt for correspondences based on cosine similarity in their paper. We follow their protocol and report the matching performance in Table~\ref{tab:cosine-matching}. As suggested by \citet{simeoni2025dinov3}, we evaluate both with and without layer normalization on the frozen features and report the best result. Interestingly, generative objectives such as \ours, MAE, and \croco~perform worse when evaluated without a linear probe. This might be related to how the DINO-family uses a patch similarity objective whereas the generative objectives tested work in pixel space. We further find that \ours~outperforms CroCo v2 and MAE and matches DINOv3 on ScanNet. 

\begin{table}[t]
\centering
     \caption{
\textbf{Cosine similarity feature matching}. We report average recall on NAVI~\cite{jampani2023navi} following the protocol of Probe3D~\cite{banani2024probing3dawarenessvisual} and EPE on MegaDepth and ScanNet. 
\label{tab:cosine-matching}} \centering
    \begin{tabular}{@{}lccc@{}}
        \toprule
        Method & NAVI~$\uparrow$ & MegaDepth~$\downarrow$ & ScanNet~$\downarrow$ \\
        \midrule
        MAE & 43.7 & 88.8 & 83.4 \\
        CroCo v2 & 41.0 & 60.3 & 62.1 \\
        DINOv3 & \bfseries 62.8 & \bfseries 36.5 & 56.4 \\
        \midrule
        \ours & 46.9 & 50.8 &  \bfseries 55.7 \\
        \bottomrule
        \end{tabular}
\end{table}

%% file: preprint.bbl
\begin{thebibliography}{81}
\providecommand{\natexlab}[1]{#1}
\providecommand{\url}[1]{\texttt{#1}}
\expandafter\ifx\csname urlstyle\endcsname\relax
  \providecommand{\doi}[1]{doi: #1}\else
  \providecommand{\doi}{doi: \begingroup \urlstyle{rm}\Url}\fi

\bibitem[An et~al.(2025)An, Kim, Park, Jung, Han, Hong, and Kim]{zeroco:cvpr}
Honggyu An, Jin~Hyeon Kim, Seonghoon Park, Jaewoo Jung, Jisang Han, Sunghwan Hong, and Seungryong Kim.
\newblock Cross-view completion models are zero-shot correspondence estimators.
\newblock In \emph{Proceedings of the IEEE/CVF Conference on Computer Vision and Pattern Recognition (CVPR)}, pages 1103--1115, 2025.

\bibitem[Assran et~al.(2023)Assran, Duval, Misra, Bojanowski, Vincent, Rabbat, LeCun, and Ballas]{ijepa:2023}
Mahmoud Assran, Quentin Duval, Ishan Misra, Piotr Bojanowski, Pascal Vincent, Michael Rabbat, Yann LeCun, and Nicolas Ballas.
\newblock Self-supervised learning from images with a joint-embedding predictive architecture.
\newblock In \emph{Proceedings of the IEEE/CVF Conference on Computer Vision and Pattern Recognition (CVPR)}, pages 15619--15629, 2023.

\bibitem[Assran et~al.(2025)Assran, Bardes, Fan, Garrido, Howes, Komeili, Muckley, Rizvi, Roberts, Sinha, Zholus, Arnaud, Gejji, Martin, Robert~Hogan, Dugas, Bojanowski, Khalidov, Labatut, Massa, Szafraniec, Krishnakumar, Li, Ma, Chandar, Meier, LeCun, Rabbat, and Ballas]{assran2025vjepa2}
Mahmoud Assran, Adrien Bardes, David Fan, Quentin Garrido, Russell Howes, Mojtaba Komeili, Matthew Muckley, Ammar Rizvi, Claire Roberts, Koustuv Sinha, Artem Zholus, Sergio Arnaud, Abha Gejji, Ada Martin, Francois Robert~Hogan, Daniel Dugas, Piotr Bojanowski, Vasil Khalidov, Patrick Labatut, Francisco Massa, Marc Szafraniec, Kapil Krishnakumar, Yong Li, Xiaodong Ma, Sarath Chandar, Franziska Meier, Yann LeCun, Michael Rabbat, and Nicolas Ballas.
\newblock V-jepa~2: Self-supervised video models enable understanding, prediction and planning.
\newblock \emph{arXiv preprint arXiv:2506.09985}, 2025.

\bibitem[Bachmann et~al.(2022)Bachmann, Mizrahi, Atanov, and Zamir]{multimae:2022}
Roman Bachmann, David Mizrahi, Andrei Atanov, and Amir Zamir.
\newblock Multimae: Multi-modal multi-task masked autoencoders, 2022.

\bibitem[Bao et~al.(2022)Bao, Dong, Piao, and Wei]{beit:2022}
Hangbo Bao, Li Dong, Songhao Piao, and Furu Wei.
\newblock Beit: Bert pre-training of image transformers, 2022.

\bibitem[Baruch et~al.(2021)Baruch, Chen, Dehghan, Dimry, Feigin, Fu, Gebauer, Joffe, Kurz, Schwartz, and Shulman]{dehghan2021arkitscenes}
Gilad Baruch, Zhuoyuan Chen, Afshin Dehghan, Tal Dimry, Yuri Feigin, Peter Fu, Thomas Gebauer, Brandon Joffe, Daniel Kurz, Arik Schwartz, and Elad Shulman.
\newblock {ARK}itscenes - a diverse real-world dataset for 3d indoor scene understanding using mobile {RGB}-d data.
\newblock In \emph{Thirty-fifth Conference on Neural Information Processing Systems Datasets and Benchmarks Track (Round 1)}, 2021.

\bibitem[Bolya et~al.(2025)Bolya, Huang, Sun, Cho, Madotto, Wei, Ma, Zhi, Rajasegaran, Rasheed, Wang, Monteiro, Xu, Dong, Ravi, Li, Doll{\'a}r, and Feichtenhofer]{bolya2025PerceptionEncoder}
Daniel Bolya, Po-Yao Huang, Peize Sun, Jang~Hyun Cho, Andrea Madotto, Chen Wei, Tengyu Ma, Jiale Zhi, Jathushan Rajasegaran, Hanoona Rasheed, Junke Wang, Marco Monteiro, Hu Xu, Shiyu Dong, Nikhila Ravi, Daniel Li, Piotr Doll{\'a}r, and Christoph Feichtenhofer.
\newblock Perception encoder: The best visual embeddings are not at the output of the network.
\newblock \emph{arXiv:2504.13181}, 2025.

\bibitem[Brynte et~al.(2024)Brynte, Iglesias, Olsson, and Kahl]{brynte-cvpr-2024}
Lucas Brynte, José Iglesias, Carl Olsson, and Fredrik Kahl.
\newblock Learning structure-from-motion with graph attention networks.
\newblock \emph{IEEE Conference on Computer Vision and Pattern Recognition}, 2024.

\bibitem[Cabon et~al.(2020)Cabon, Murray, and Humenberger]{cabon2020vkitti2}
Yohann Cabon, Naila Murray, and Martin Humenberger.
\newblock Virtual kitti 2, 2020.

\bibitem[Cabon et~al.(2025{\natexlab{a}})Cabon, Stoffl, Antsfeld, Csurka, Chidlovskii, Revaud, and Leroy]{mast3r:2024}
Yohann Cabon, Lucas Stoffl, Leonid Antsfeld, Gabriela Csurka, Boris Chidlovskii, Jerome Revaud, and Vincent Leroy.
\newblock Must3r: Multi-view network for stereo 3d reconstruction.
\newblock In \emph{Proceedings of the IEEE/CVF Conference on Computer Vision and Pattern Recognition (CVPR)}, pages 1050--1060, 2025{\natexlab{a}}.

\bibitem[Cabon et~al.(2025{\natexlab{b}})Cabon, Stoffl, Antsfeld, Csurka, Chidlovskii, Revaud, and Leroy]{must3r:2025}
Yohann Cabon, Lucas Stoffl, Leonid Antsfeld, Gabriela Csurka, Boris Chidlovskii, Jerome Revaud, and Vincent Leroy.
\newblock Must3r: Multi-view network for stereo 3d reconstruction.
\newblock In \emph{Proceedings of the IEEE/CVF Conference on Computer Vision and Pattern Recognition (CVPR)}, pages 1050--1060, 2025{\natexlab{b}}.

\bibitem[Caron et~al.(2021)Caron, Touvron, Misra, Jégou, Mairal, Bojanowski, and Joulin]{dino}
Mathilde Caron, Hugo Touvron, Ishan Misra, Hervé Jégou, Julien Mairal, Piotr Bojanowski, and Armand Joulin.
\newblock Emerging properties in self-supervised vision transformers, 2021.

\bibitem[Carreira et~al.(2025)Carreira, Gokay, King, Zhang, Rocco, Mahendran, Keck, Heyward, Koppula, Pot, Erdogan, Hasson, Yang, Greff, Moing, van Steenkiste, Zoran, Hudson, Vélez, Polanía, Friedman, Duvarney, Goroshin, Allen, Walker, Kabra, Aboussouan, Sun, Kipf, Doersch, Pătrăucean, Damen, Luc, Sajjadi, and Zisserman]{carreira2025scaling4drepresentations}
João Carreira, Dilara Gokay, Michael King, Chuhan Zhang, Ignacio Rocco, Aravindh Mahendran, Thomas~Albert Keck, Joseph Heyward, Skanda Koppula, Etienne Pot, Goker Erdogan, Yana Hasson, Yi Yang, Klaus Greff, Guillaume~Le Moing, Sjoerd van Steenkiste, Daniel Zoran, Drew~A. Hudson, Pedro Vélez, Luisa Polanía, Luke Friedman, Chris Duvarney, Ross Goroshin, Kelsey Allen, Jacob Walker, Rishabh Kabra, Eric Aboussouan, Jennifer Sun, Thomas Kipf, Carl Doersch, Viorica Pătrăucean, Dima Damen, Pauline Luc, Mehdi S.~M. Sajjadi, and Andrew Zisserman.
\newblock Scaling 4d representations, 2025.

\bibitem[Contributors(2018)]{mmcv}
MMCV Contributors.
\newblock {MMCV: OpenMMLab} computer vision foundation.
\newblock \url{https://github.com/open-mmlab/mmcv}, 2018.

\bibitem[Dai et~al.(2017)Dai, Chang, Savva, Halber, Funkhouser, and Nie{\ss}ner]{dai2017scannet}
Angela Dai, Angel~X Chang, Manolis Savva, Maciej Halber, Thomas Funkhouser, and Matthias Nie{\ss}ner.
\newblock Scannet: Richly-annotated 3d reconstructions of indoor scenes.
\newblock In \emph{Proceedings of the IEEE conference on computer vision and pattern recognition}, pages 5828--5839, 2017.

\bibitem[Darcet et~al.(2025)Darcet, Baldassarre, Oquab, Mairal, and Bojanowski]{CAPI:2025}
Timothée Darcet, Federico Baldassarre, Maxime Oquab, Julien Mairal, and Piotr Bojanowski.
\newblock Cluster and predict latent patches for improved masked image modeling, 2025.

\bibitem[Deng et~al.(2009)Deng, Dong, Socher, Li, Li, and Fei-Fei]{imageNet2009}
Jia Deng, Wei Dong, Richard Socher, Li-Jia Li, Kai Li, and Li Fei-Fei.
\newblock Imagenet: A large-scale hierarchical image database.
\newblock In \emph{2009 IEEE Conference on Computer Vision and Pattern Recognition}, pages 248--255, 2009.

\bibitem[Devlin et~al.(2019)Devlin, Chang, Lee, and Toutanova]{BERT:2019}
Jacob Devlin, Ming-Wei Chang, Kenton Lee, and Kristina Toutanova.
\newblock {BERT}: Pre-training of deep bidirectional transformers for language understanding.
\newblock In \emph{Proceedings of the 2019 Conference of the North {A}merican Chapter of the Association for Computational Linguistics: Human Language Technologies, Volume 1 (Long and Short Papers)}, pages 4171--4186, Minneapolis, Minnesota, 2019. Association for Computational Linguistics.

\bibitem[Dong et~al.(2025)Dong, Wang, Liu, Cai, Fan, Kannala, and Yang]{reloc3r:2025}
Siyan Dong, Shuzhe Wang, Shaohui Liu, Lulu Cai, Qingnan Fan, Juho Kannala, and Yanchao Yang.
\newblock Reloc3r: Large-scale training of relative camera pose regression for generalizable, fast, and accurate visual localization.
\newblock In \emph{Proceedings of the IEEE/CVF Conference on Computer Vision and Pattern Recognition (CVPR)}, pages 16739--16752, 2025.

\bibitem[Dosovitskiy et~al.(2021)Dosovitskiy, Beyer, Kolesnikov, Weissenborn, Zhai, Unterthiner, Dehghani, Minderer, Heigold, Gelly, Uszkoreit, and Houlsby]{dosovitskiy2021an}
Alexey Dosovitskiy, Lucas Beyer, Alexander Kolesnikov, Dirk Weissenborn, Xiaohua Zhai, Thomas Unterthiner, Mostafa Dehghani, Matthias Minderer, Georg Heigold, Sylvain Gelly, Jakob Uszkoreit, and Neil Houlsby.
\newblock An image is worth 16x16 words: Transformers for image recognition at scale.
\newblock In \emph{International Conference on Learning Representations}, 2021.

\bibitem[Dusmanu et~al.(2019)Dusmanu, Rocco, Pajdla, Pollefeys, Sivic, Torii, and Sattler]{d2net:2019}
Mihai Dusmanu, Ignacio Rocco, Tomas Pajdla, Marc Pollefeys, Josef Sivic, Akihiko Torii, and Torsten Sattler.
\newblock {D2-Net: A Trainable CNN for Joint Detection and Description of Local Features}.
\newblock In \emph{Proceedings of the 2019 IEEE/CVF Conference on Computer Vision and Pattern Recognition}, 2019.

\bibitem[Edstedt et~al.(2023)Edstedt, Athanasiadis, Wadenb\"ack, and Felsberg]{edstedt2022dkmdensekernelizedfeature}
Johan Edstedt, Ioannis Athanasiadis, M\r{a}rten Wadenb\"ack, and Michael Felsberg.
\newblock Dkm: Dense kernelized feature matching for geometry estimation.
\newblock In \emph{Proceedings of the IEEE/CVF Conference on Computer Vision and Pattern Recognition (CVPR)}, pages 17765--17775, 2023.

\bibitem[Edstedt et~al.(2024)Edstedt, Sun, Bökman, Wadenbäck, and Felsberg]{roma:2023}
Johan Edstedt, Qiyu Sun, Georg Bökman, Mårten Wadenbäck, and Michael Felsberg.
\newblock {RoMa: Robust Dense Feature Matching}.
\newblock \emph{IEEE Conference on Computer Vision and Pattern Recognition}, 2024.

\bibitem[El~Banani et~al.(2024)El~Banani, Raj, Maninis, Kar, Li, Rubinstein, Sun, Guibas, Johnson, and Jampani]{banani2024probing3dawarenessvisual}
Mohamed El~Banani, Amit Raj, Kevis-Kokitsi Maninis, Abhishek Kar, Yuanzhen Li, Michael Rubinstein, Deqing Sun, Leonidas Guibas, Justin Johnson, and Varun Jampani.
\newblock Probing the 3d awareness of visual foundation models.
\newblock In \emph{Proceedings of the IEEE/CVF Conference on Computer Vision and Pattern Recognition (CVPR)}, pages 21795--21806, 2024.

\bibitem[El-Nouby et~al.(2024)El-Nouby, Klein, Zhai, Bautista, Toshev, Shankar, Susskind, and Joulin]{AIM:2024}
Alaaeldin El-Nouby, Michal Klein, Shuangfei Zhai, Miguel~Angel Bautista, Alexander Toshev, Vaishaal Shankar, Joshua~M Susskind, and Armand Joulin.
\newblock Scalable pre-training of large autoregressive image models.
\newblock In \emph{Int. Conf. Machine Learning}, 2024.

\bibitem[Fang et~al.(2023)Fang, Wang, Xie, Sun, Wu, Wang, Huang, Wang, and Cao]{eva:2022}
Yuxin Fang, Wen Wang, Binhui Xie, Quan Sun, Ledell Wu, Xinggang Wang, Tiejun Huang, Xinlong Wang, and Yue Cao.
\newblock Eva: Exploring the limits of masked visual representation learning at scale.
\newblock In \emph{Proceedings of the IEEE/CVF Conference on Computer Vision and Pattern Recognition (CVPR)}, pages 19358--19369, 2023.

\bibitem[Goyal et~al.(2018)Goyal, Dollár, Girshick, Noordhuis, Wesolowski, Kyrola, Tulloch, Jia, and He]{goyal2018accuratelargeminibatchsgd}
Priya Goyal, Piotr Dollár, Ross Girshick, Pieter Noordhuis, Lukasz Wesolowski, Aapo Kyrola, Andrew Tulloch, Yangqing Jia, and Kaiming He.
\newblock Accurate, large minibatch sgd: Training imagenet in 1 hour, 2018.

\bibitem[Hartley and Zisserman(2004)]{Hartley_Zisserman_2004}
Richard Hartley and Andrew Zisserman.
\newblock \emph{Multiple View Geometry in Computer Vision}.
\newblock Cambridge University Press, 2 edition, 2004.

\bibitem[He et~al.(2022)He, Chen, Xie, Li, Doll\'ar, and Girshick]{MAE:2021}
Kaiming He, Xinlei Chen, Saining Xie, Yanghao Li, Piotr Doll\'ar, and Ross Girshick.
\newblock Masked autoencoders are scalable vision learners.
\newblock In \emph{Proceedings of the IEEE/CVF Conference on Computer Vision and Pattern Recognition (CVPR)}, pages 16000--16009, 2022.

\bibitem[Huang et~al.(2018)Huang, Matzen, Kopf, Ahuja, and Huang]{DeepMVS}
Po-Han Huang, Kevin Matzen, Johannes Kopf, Narendra Ahuja, and Jia-Bin Huang.
\newblock Deepmvs: Learning multi-view stereopsis.
\newblock In \emph{IEEE Conference on Computer Vision and Pattern Recognition (CVPR)}, 2018.

\bibitem[Jampani et~al.(2023)Jampani, Maninis, Engelhardt, Karpur, Truong, Sargent, Popov, Araujo, Martin-Brualla, Patel, Vlasic, Ferrari, Makadia, Liu, Li, and Zhou]{jampani2023navi}
Varun Jampani, Kevis-Kokitsi Maninis, Andreas Engelhardt, Arjun Karpur, Karen Truong, Kyle Sargent, Stefan Popov, Andre Araujo, Ricardo Martin-Brualla, Kaushal Patel, Daniel Vlasic, Vittorio Ferrari, Ameesh Makadia, Ce Liu, Yuanzhen Li, and Howard Zhou.
\newblock Navi: Category-agnostic image collections with high-quality 3d shape and pose annotations.
\newblock In \emph{NeurIPS}, 2023.

\bibitem[Jensen et~al.(2014)Jensen, Dahl, Vogiatzis, Tola, and Aanaes]{dtu:2014}
Rasmus Jensen, Anders Dahl, George Vogiatzis, Engin Tola, and Henrik Aanaes.
\newblock Large scale multi-view stereopsis evaluation.
\newblock In \emph{Proceedings of the IEEE Conference on Computer Vision and Pattern Recognition (CVPR)}, 2014.

\bibitem[Keetha et~al.(2025)Keetha, M\"{u}ller, Sch\"{o}nberger, Porzi, Zhang, Fischer, Knapitsch, Zauss, Weber, Antunes, Luiten, Lopez-Antequera, Bul\`{o}, Richardt, Ramanan, Scherer, and Kontschieder]{keetha2025mapanything}
Nikhil Keetha, Norman M\"{u}ller, Johannes Sch\"{o}nberger, Lorenzo Porzi, Yuchen Zhang, Tobias Fischer, Arno Knapitsch, Duncan Zauss, Ethan Weber, Nelson Antunes, Jonathon Luiten, Manuel Lopez-Antequera, Samuel~Rota Bul\`{o}, Christian Richardt, Deva Ramanan, Sebastian Scherer, and Peter Kontschieder.
\newblock {MapAnything}: Universal feed-forward metric {3D} reconstruction, 2025.
\newblock arXiv preprint arXiv:2509.13414.

\bibitem[Li and Snavely(2018)]{MegaDepthLi18}
Zhengqi Li and Noah Snavely.
\newblock Megadepth: Learning single-view depth prediction from internet photos.
\newblock In \emph{Computer Vision and Pattern Recognition (CVPR)}, 2018.

\bibitem[Lindenberger et~al.(2023)Lindenberger, Sarlin, and Pollefeys]{lightglue:2023}
Philipp Lindenberger, Paul-Edouard Sarlin, and Marc Pollefeys.
\newblock Lightglue: Local feature matching at light speed.
\newblock In \emph{Proceedings of the IEEE/CVF International Conference on Computer Vision (ICCV)}, pages 17627--17638, 2023.

\bibitem[Ling et~al.(2024)Ling, Sheng, Tu, Zhao, Xin, Wan, Yu, Guo, Yu, Lu, et~al.]{ling2024dl3dv}
Lu Ling, Yichen Sheng, Zhi Tu, Wentian Zhao, Cheng Xin, Kun Wan, Lantao Yu, Qianyu Guo, Zixun Yu, Yawen Lu, et~al.
\newblock Dl3dv-10k: A large-scale scene dataset for deep learning-based 3d vision.
\newblock In \emph{Proceedings of the IEEE/CVF Conference on Computer Vision and Pattern Recognition}, pages 22160--22169, 2024.

\bibitem[Loiseau et~al.(2025)Loiseau, Bourmaud, and Lepetit]{aligat0r:2025}
Thibaut Loiseau, Guillaume Bourmaud, and Vincent Lepetit.
\newblock Alligat0r: Pre-training through co-visibility segmentation for relative camera pose regression, 2025.

\bibitem[Loshchilov and Hutter(2017)]{adamw:2019}
Ilya Loshchilov and Frank Hutter.
\newblock Decoupled weight decay regularization.
\newblock In \emph{International Conference on Learning Representations}, 2017.

\bibitem[Nathan~Silberman and Fergus(2012)]{NYUd:2012}
Pushmeet~Kohli Nathan~Silberman, Derek~Hoiem and Rob Fergus.
\newblock Indoor segmentation and support inference from rgbd images.
\newblock In \emph{ECCV}, 2012.

\bibitem[Oquab et~al.(2024)Oquab, Darcet, Moutakanni, Vo, Szafraniec, Khalidov, Fernandez, Haziza, Massa, El-Nouby, Assran, Ballas, Galuba, Howes, Huang, Li, Misra, Rabbat, Sharma, Synnaeve, Xu, Jegou, Mairal, Labatut, Joulin, and Bojanowski]{dinov2}
Maxime Oquab, Timothée Darcet, Théo Moutakanni, Huy Vo, Marc Szafraniec, Vasil Khalidov, Pierre Fernandez, Daniel Haziza, Francisco Massa, Alaaeldin El-Nouby, Mahmoud Assran, Nicolas Ballas, Wojciech Galuba, Russell Howes, Po-Yao Huang, Shang-Wen Li, Ishan Misra, Michael Rabbat, Vasu Sharma, Gabriel Synnaeve, Hu Xu, Hervé Jegou, Julien Mairal, Patrick Labatut, Armand Joulin, and Piotr Bojanowski.
\newblock Dinov2: Learning robust visual features without supervision.
\newblock \emph{Transactions on Machine Learning Research}, 2024.

\bibitem[Radford et~al.(2019)Radford, Wu, Child, Luan, Amodei, and Sutskever]{gpt2:2019}
Alec Radford, Jeffrey Wu, Rewon Child, David Luan, Dario Amodei, and Ilya Sutskever.
\newblock Language models are unsupervised multitask learners.
\newblock \emph{OpenAI}, 2019.
\newblock Accessed: 2024-11-15.

\bibitem[Ranftl et~al.(2021)Ranftl, Bochkovskiy, and Koltun]{ranftl2021vision}
Ren\'e Ranftl, Alexey Bochkovskiy, and Vladlen Koltun.
\newblock Vision transformers for dense prediction.
\newblock In \emph{Proceedings of the IEEE/CVF International Conference on Computer Vision (ICCV)}, pages 12179--12188, 2021.

\bibitem[Reizenstein et~al.(2021)Reizenstein, Shapovalov, Henzler, Sbordone, Labatut, and Novotny]{co3d:2021}
Jeremy Reizenstein, Roman Shapovalov, Philipp Henzler, Luca Sbordone, Patrick Labatut, and David Novotny.
\newblock Common objects in 3d: Large-scale learning and evaluation of real-life 3d category reconstruction.
\newblock In \emph{International Conference on Computer Vision}, 2021.

\bibitem[Roberts et~al.(2021)Roberts, Ramapuram, Ranjan, Kumar, Bautista, Paczan, Webb, and Susskind]{hypersim:2021}
Mike Roberts, Jason Ramapuram, Anurag Ranjan, Atulit Kumar, Miguel~Angel Bautista, Nathan Paczan, Russ Webb, and Joshua~M. Susskind.
\newblock {Hypersim}: {A} photorealistic synthetic dataset for holistic indoor scene understanding.
\newblock In \emph{International Conference on Computer Vision (ICCV) 2021}, 2021.

\bibitem[Russakovsky et~al.(2015)Russakovsky, Deng, Su, Krause, Satheesh, Ma, Huang, Karpathy, Khosla, Bernstein, et~al.]{russakovsky2015imagenet}
Olga Russakovsky, Jia Deng, Hao Su, Jonathan Krause, Sanjeev Satheesh, Sean Ma, Zhiheng Huang, Andrej Karpathy, Aditya Khosla, Michael Bernstein, et~al.
\newblock Imagenet large scale visual recognition challenge.
\newblock \emph{International journal of computer vision}, 115:\penalty0 211--252, 2015.

\bibitem[Sarlin et~al.(2020)Sarlin, DeTone, Malisiewicz, and Rabinovich]{sarlin2020superglue}
Paul-Edouard Sarlin, Daniel DeTone, Tomasz Malisiewicz, and Andrew Rabinovich.
\newblock Superglue: Learning feature matching with graph neural networks.
\newblock In \emph{Proceedings of the IEEE/CVF conference on computer vision and pattern recognition}, pages 4938--4947, 2020.

\bibitem[Schops et~al.(2017)Schops, Schonberger, Galliani, Sattler, Schindler, Pollefeys, and Geiger]{eth3d:2017}
Thomas Schops, Johannes~L. Schonberger, Silvano Galliani, Torsten Sattler, Konrad Schindler, Marc Pollefeys, and Andreas Geiger.
\newblock A multi-view stereo benchmark with high-resolution images and multi-camera videos.
\newblock In \emph{Proceedings of the IEEE Conference on Computer Vision and Pattern Recognition (CVPR)}, 2017.

\bibitem[Seo et~al.(2023)Seo, Kim, James, Lee, Shin, and Abbeel]{mv-worldmodels}
Younggyo Seo, Junsu Kim, Stephen James, Kimin Lee, Jinwoo Shin, and Pieter Abbeel.
\newblock Multi-view masked world models for visual robotic manipulation.
\newblock In \emph{Proceedings of the 40th International Conference on Machine Learning}, pages 30613--30632. PMLR, 2023.

\bibitem[Shah et~al.(2024)Shah, Crandall, Xu, Zhou, George, Bansal, and Chellappa]{mv2mae}
Ketul Shah, Robert Crandall, Jie Xu, Peng Zhou, Marian George, Mayank Bansal, and Rama Chellappa.
\newblock Mv2mae: Multi-view video masked autoencoders, 2024.

\bibitem[Sim{\'e}oni et~al.(2025)Sim{\'e}oni, Vo, Seitzer, Baldassarre, Oquab, Jose, Khalidov, Szafraniec, Yi, Ramamonjisoa, Massa, Haziza, Wehrstedt, Wang, Darcet, Moutakanni, Sentana, Roberts, Vedaldi, Tolan, Brandt, Couprie, Mairal, J{\'e}gou, Labatut, and Bojanowski]{simeoni2025dinov3}
Oriane Sim{\'e}oni, Huy~V. Vo, Maximilian Seitzer, Federico Baldassarre, Maxime Oquab, Cijo Jose, Vasil Khalidov, Marc Szafraniec, Seungeun Yi, Micha{\"e}l Ramamonjisoa, Francisco Massa, Daniel Haziza, Luca Wehrstedt, Jianyuan Wang, Timoth{\'e}e Darcet, Th{\'e}o Moutakanni, Leonel Sentana, Claire Roberts, Andrea Vedaldi, Jamie Tolan, John Brandt, Camille Couprie, Julien Mairal, Herv{\'e} J{\'e}gou, Patrick Labatut, and Piotr Bojanowski.
\newblock {DINOv3}, 2025.

\bibitem[Su et~al.(2023)Su, Lu, Pan, Murtadha, Wen, and Liu]{rope}
Jianlin Su, Yu Lu, Shengfeng Pan, Ahmed Murtadha, Bo Wen, and Yunfeng Liu.
\newblock Roformer: Enhanced transformer with rotary position embedding, 2023.

\bibitem[Sun et~al.(2021{\natexlab{a}})Sun, Shen, Wang, Bao, and Zhou]{loftr:2021}
Jiaming Sun, Zehong Shen, Yuang Wang, Hujun Bao, and Xiaowei Zhou.
\newblock Loftr: Detector-free local feature matching with transformers.
\newblock \emph{CoRR}, abs/2104.00680, 2021{\natexlab{a}}.

\bibitem[Sun et~al.(2021{\natexlab{b}})Sun, Shen, Wang, Bao, and Zhou]{sun2021loftr}
Jiaming Sun, Zehong Shen, Yuang Wang, Hujun Bao, and Xiaowei Zhou.
\newblock {LoFTR: Detector-free local feature matching with transformers}.
\newblock In \emph{Proceedings of the IEEE/CVF Conference on Computer Vision and Pattern Recognition}, pages 8922--8931, 2021{\natexlab{b}}.

\bibitem[Tang et~al.(2025)Tang, Fan, Wang, Xu, Ranjan, Schwing, and Yan]{mvdust3r:2024}
Zhenggang Tang, Yuchen Fan, Dilin Wang, Hongyu Xu, Rakesh Ranjan, Alexander Schwing, and Zhicheng Yan.
\newblock Mv-dust3r+: Single-stage scene reconstruction from sparse views in 2 seconds.
\newblock In \emph{Proceedings of the IEEE/CVF Conference on Computer Vision and Pattern Recognition (CVPR)}, pages 5283--5293, 2025.

\bibitem[Tarvainen and Valpola(2017)]{meanteachers:2017}
Antti Tarvainen and Harri Valpola.
\newblock Mean teachers are better role models: Weight-averaged consistency targets improve semi-supervised deep learning results.
\newblock In \emph{Advances in Neural Information Processing Systems}. Curran Associates, Inc., 2017.

\bibitem[Tong et~al.(2022)Tong, Song, Wang, and Wang]{tong2022videomae}
Zhan Tong, Yibing Song, Jue Wang, and Limin Wang.
\newblock Video{MAE}: Masked autoencoders are data-efficient learners for self-supervised video pre-training.
\newblock In \emph{Advances in Neural Information Processing Systems}, 2022.

\bibitem[Touvron et~al.(2023)Touvron, Martin, Stone, Albert, Almahairi, Babaei, Bashlykov, Batra, Bhargava, Bhosale, Bikel, Blecher, Ferrer, Chen, Cucurull, Esiobu, Fernandes, Fu, Fu, Fuller, Gao, Goswami, Goyal, Hartshorn, Hosseini, Hou, Inan, Kardas, Kerkez, Khabsa, Kloumann, Korenev, Koura, Lachaux, Lavril, Lee, Liskovich, Lu, Mao, Martinet, Mihaylov, Mishra, Molybog, Nie, Poulton, Reizenstein, Rungta, Saladi, Schelten, Silva, Smith, Subramanian, Tan, Tang, Taylor, Williams, Kuan, Xu, Yan, Zarov, Zhang, Fan, Kambadur, Narang, Rodriguez, Stojnic, Edunov, and Scialom]{llama2}
Hugo Touvron, Louis Martin, Kevin Stone, Peter Albert, Amjad Almahairi, Yasmine Babaei, Nikolay Bashlykov, Soumya Batra, Prajjwal Bhargava, Shruti Bhosale, Dan Bikel, Lukas Blecher, Cristian~Canton Ferrer, Moya Chen, Guillem Cucurull, David Esiobu, Jude Fernandes, Jeremy Fu, Wenyin Fu, Brian Fuller, Cynthia Gao, Vedanuj Goswami, Naman Goyal, Anthony Hartshorn, Saghar Hosseini, Rui Hou, Hakan Inan, Marcin Kardas, Viktor Kerkez, Madian Khabsa, Isabel Kloumann, Artem Korenev, Punit~Singh Koura, Marie-Anne Lachaux, Thibaut Lavril, Jenya Lee, Diana Liskovich, Yinghai Lu, Yuning Mao, Xavier Martinet, Todor Mihaylov, Pushkar Mishra, Igor Molybog, Yixin Nie, Andrew Poulton, Jeremy Reizenstein, Rashi Rungta, Kalyan Saladi, Alan Schelten, Ruan Silva, Eric~Michael Smith, Ranjan Subramanian, Xiaoqing~Ellen Tan, Binh Tang, Ross Taylor, Adina Williams, Jian~Xiang Kuan, Puxin Xu, Zheng Yan, Iliyan Zarov, Yuchen Zhang, Angela Fan, Melanie Kambadur, Sharan Narang, Aurelien Rodriguez, Robert Stojnic, Sergey Edunov, and Thomas
  Scialom.
\newblock Llama 2: Open foundation and fine-tuned chat models, 2023.

\bibitem[Tschannen et~al.(2025)Tschannen, Gritsenko, Wang, Naeem, Alabdulmohsin, Parthasarathy, Evans, Beyer, Xia, Mustafa, Hénaff, Harmsen, Steiner, and Zhai]{siglip2}
Michael Tschannen, Alexey Gritsenko, Xiao Wang, Muhammad~Ferjad Naeem, Ibrahim Alabdulmohsin, Nikhil Parthasarathy, Talfan Evans, Lucas Beyer, Ye Xia, Basil Mustafa, Olivier Hénaff, Jeremiah Harmsen, Andreas Steiner, and Xiaohua Zhai.
\newblock Siglip 2: Multilingual vision-language encoders with improved semantic understanding, localization, and dense features, 2025.

\bibitem[Tung et~al.(2024)Tung, Chou, Cai, Yang, Zhang, Wetzstein, Hariharan, and Snavely]{tung2024megascenes}
Joseph Tung, Gene Chou, Ruojin Cai, Guandao Yang, Kai Zhang, Gordon Wetzstein, Bharath Hariharan, and Noah Snavely.
\newblock Megascenes: Scene-level view synthesis at scale.
\newblock In \emph{ECCV}, 2024.

\bibitem[Uhrig et~al.(2017)Uhrig, Schneider, Schneider, Franke, Brox, and Geiger]{KITTI:2017}
Jonas Uhrig, Nick Schneider, Lukas Schneider, Uwe Franke, Thomas Brox, and Andreas Geiger.
\newblock Sparsity invariant cnns.
\newblock In \emph{International Conference on 3D Vision (3DV)}, 2017.

\bibitem[Wang and Agapito(2024)]{spann3r:2024}
Hengyi Wang and Lourdes Agapito.
\newblock 3d reconstruction with spatial memory, 2024.

\bibitem[Wang et~al.(2025{\natexlab{a}})Wang, Chen, Karaev, Vedaldi, Rupprecht, and Novotny]{vggt:2025}
Jianyuan Wang, Minghao Chen, Nikita Karaev, Andrea Vedaldi, Christian Rupprecht, and David Novotny.
\newblock Vggt: Visual geometry grounded transformer, 2025{\natexlab{a}}.

\bibitem[Wang et~al.(2023)Wang, Huang, Zhao, Tong, He, Wang, Wang, and Qiao]{wang2023videomaev2}
Limin Wang, Bingkun Huang, Zhiyu Zhao, Zhan Tong, Yinan He, Yi Wang, Yali Wang, and Yu Qiao.
\newblock Videomae v2: Scaling video masked autoencoders with dual masking.
\newblock In \emph{Proceedings of the IEEE/CVF Conference on Computer Vision and Pattern Recognition (CVPR)}, pages 14549--14560, 2023.

\bibitem[Wang et~al.(2024)Wang, Leroy, Cabon, Chidlovskii, and Revaud]{dust3r:2024}
Shuzhe Wang, Vincent Leroy, Yohann Cabon, Boris Chidlovskii, and Jerome Revaud.
\newblock Dust3r: Geometric 3d vision made easy, 2024.

\bibitem[Wang et~al.(2025{\natexlab{b}})Wang, Zhou, Zhu, Chang, Zhou, Li, Chen, Pang, Shen, and He]{wang2025pi3}
Yifan Wang, Jianjun Zhou, Haoyi Zhu, Wenzheng Chang, Yang Zhou, Zizun Li, Junyi Chen, Jiangmiao Pang, Chunhua Shen, and Tong He.
\newblock $\pi^3$: Scalable permutation-equivariant visual geometry learning, 2025{\natexlab{b}}.

\bibitem[Weinzaepfel et~al.(2022)Weinzaepfel, Leroy, Lucas, Br{\'e}gier, Cabon, Arora, Antsfeld, Chidlovskii, Csurka, and Revaud]{croco}
Philippe Weinzaepfel, Vincent Leroy, Thomas Lucas, Romain Br{\'e}gier, Yohann Cabon, Vaibhav Arora, Leonid Antsfeld, Boris Chidlovskii, Gabriela Csurka, and J{\'e}r{\^o}me Revaud.
\newblock Croco: Self-supervised pre-training for 3d vision tasks by cross-view completion.
\newblock \emph{Advances in Neural Information Processing Systems}, 35:\penalty0 3502--3516, 2022.

\bibitem[Weinzaepfel et~al.(2023)Weinzaepfel, Lucas, Leroy, Cabon, Arora, Br{\'e}gier, Csurka, Antsfeld, Chidlovskii, and Revaud]{crocov2}
Philippe Weinzaepfel, Thomas Lucas, Vincent Leroy, Yohann Cabon, Vaibhav Arora, Romain Br{\'e}gier, Gabriela Csurka, Leonid Antsfeld, Boris Chidlovskii, and J{\'e}r{\^o}me Revaud.
\newblock Croco v2: Improved cross-view completion pre-training for stereo matching and optical flow.
\newblock In \emph{Proceedings of the IEEE/CVF International Conference on Computer Vision}, pages 17969--17980, 2023.

\bibitem[Wu et~al.(2025)Wu, Zhang, Pai, Wang, Singh, Yang, Gao, and Ma]{wu2025simdino}
Ziyang Wu, Jingyuan Zhang, Druv Pai, Xudong Wang, Chandan Singh, Jianwei Yang, Jianfeng Gao, and Yi Ma.
\newblock Simplifying dino via coding rate regularization.
\newblock In \emph{Int. Conf. Machine Learning}, 2025.

\bibitem[Xia et~al.(2024)Xia, Fu, Liu, and Wang]{wilrgbd:2024}
Hongchi Xia, Yang Fu, Sifei Liu, and Xiaolong Wang.
\newblock Rgbd objects in the wild: Scaling real-world 3d object learning from rgb-d videos.
\newblock In \emph{Proceedings of the IEEE/CVF Conference on Computer Vision and Pattern Recognition (CVPR)}, 2024.

\bibitem[Xie et~al.(2022)Xie, Zhang, Cao, Lin, Bao, Yao, Dai, and Hu]{xie2022simmimsimpleframeworkmasked}
Zhenda Xie, Zheng Zhang, Yue Cao, Yutong Lin, Jianmin Bao, Zhuliang Yao, Qi Dai, and Han Hu.
\newblock Simmim: A simple framework for masked image modeling.
\newblock In \emph{Proceedings of the IEEE/CVF Conference on Computer Vision and Pattern Recognition (CVPR)}, pages 9653--9663, 2022.

\bibitem[Yang et~al.(2025)Yang, Sax, Liang, Henaff, Tang, Cao, Chai, Meier, and Feiszli]{fast3r:2025}
Jianing Yang, Alexander Sax, Kevin~J. Liang, Mikael Henaff, Hao Tang, Ang Cao, Joyce Chai, Franziska Meier, and Matt Feiszli.
\newblock Fast3r: Towards 3d reconstruction of 1000+ images in one forward pass.
\newblock In \emph{Proceedings of the IEEE/CVF Conference on Computer Vision and Pattern Recognition (CVPR)}, pages 21924--21935, 2025.

\bibitem[Yao et~al.(2020)Yao, Luo, Li, Zhang, Ren, Zhou, Fang, and Quan]{yao2020blendedmvs}
Yao Yao, Zixin Luo, Shiwei Li, Jingyang Zhang, Yufan Ren, Lei Zhou, Tian Fang, and Long Quan.
\newblock Blendedmvs: A large-scale dataset for generalized multi-view stereo networks.
\newblock \emph{Computer Vision and Pattern Recognition (CVPR)}, 2020.

\bibitem[Yeshwanth et~al.(2023)Yeshwanth, Liu, Nie{\ss}ner, and Dai]{scannetpp:2023}
Chandan Yeshwanth, Yueh-Cheng Liu, Matthias Nie{\ss}ner, and Angela Dai.
\newblock Scannet++: A high-fidelity dataset of 3d indoor scenes.
\newblock In \emph{Proceedings of the International Conference on Computer Vision ({ICCV})}, 2023.

\bibitem[Yugay et~al.(2025)Yugay, Nguyen, Gevers, Snoek, and Oswald]{yugay2025visualodometrytransformers}
Vlardimir Yugay, Duy-Kien Nguyen, Theo Gevers, Cees G.~M. Snoek, and Martin~R. Oswald.
\newblock Visual odometry with transformers, 2025.

\bibitem[Zamir et~al.(2016)Zamir, Wekel, Agrawal, Wei, Malik, and Savarese]{streetview:2016}
Amir~R Zamir, Tilman Wekel, Pulkit Agrawal, Colin Wei, Jitendra Malik, and Silvio Savarese.
\newblock Generic {3D} representation via pose estimation and matching.
\newblock In \emph{European Conference on Computer Vision}, pages 535--553. Springer, 2016.

\bibitem[Zhang et~al.(2025)Zhang, Keetha, Lyu, Jhamb, Chen, Qiu, Karhade, Jha, Hu, Ramanan, Scherer, and Wang]{zhang2025ufm}
Yuchen Zhang, Nikhil Keetha, Chenwei Lyu, Bhuvan Jhamb, Yutian Chen, Yuheng Qiu, Jay Karhade, Shreyas Jha, Yaoyu Hu, Deva Ramanan, Sebastian Scherer, and Wenshan Wang.
\newblock Ufm: A simple path towards unified dense correspondence with flow.
\newblock In \emph{neurips}, 2025.

\bibitem[Zheng et~al.(2023)Zheng, Harley, Shen, Wetzstein, and Guibas]{zheng2023point}
Yang Zheng, Adam~W. Harley, Bokui Shen, Gordon Wetzstein, and Leonidas~J. Guibas.
\newblock Pointodyssey: A large-scale synthetic dataset for long-term point tracking.
\newblock In \emph{ICCV}, 2023.

\bibitem[Zhou et~al.(2017)Zhou, Zhao, Puig, Fidler, Barriuso, and Torralba]{ade20k:2017}
Bolei Zhou, Hang Zhao, Xavier Puig, Sanja Fidler, Adela Barriuso, and Antonio Torralba.
\newblock Scene parsing through ade20k dataset.
\newblock In \emph{Proceedings of the IEEE Conference on Computer Vision and Pattern Recognition}, 2017.

\bibitem[Zhou et~al.(2019)Zhou, Zhao, Puig, Xiao, Fidler, Barriuso, and Torralba]{ade20k:2019}
Bolei Zhou, Hang Zhao, Xavier Puig, Tete Xiao, Sanja Fidler, Adela Barriuso, and Antonio Torralba.
\newblock Semantic understanding of scenes through the ade20k dataset.
\newblock \emph{International Journal of Computer Vision}, 127\penalty0 (3):\penalty0 302--321, 2019.

\bibitem[Zhou et~al.(2022)Zhou, Wei, Wang, Shen, Xie, Yuille, and Kong]{iBOT:2022}
Jinghao Zhou, Chen Wei, Huiyu Wang, Wei Shen, Cihang Xie, Alan Yuille, and Tao Kong.
\newblock ibot: Image bert pre-training with online tokenizer.
\newblock \emph{International Conference on Learning Representations (ICLR)}, 2022.

\bibitem[Zhou et~al.(2018)Zhou, Tucker, Flynn, Fyffe, and Snavely]{realestate10k}
Tinghui Zhou, Richard Tucker, John Flynn, Graham Fyffe, and Noah Snavely.
\newblock Stereo magnification: Learning view synthesis using multiplane images.
\newblock \emph{ACM Trans. Graph. (Proc. SIGGRAPH)}, 37, 2018.

\end{thebibliography}
